\pdfoutput=1

\documentclass[11pt]{article}

\usepackage[preprint]{acl}

\usepackage{times}
\usepackage{latexsym}
\usepackage{booktabs}
\usepackage[T1]{fontenc}

\usepackage[utf8]{inputenc}

\usepackage{microtype}

\usepackage{inconsolata}

\usepackage{graphicx}

\usepackage{bbm}

\usepackage{enumitem}

\usepackage{siunitx}
\usepackage{caption}
\usepackage{multirow}

\usepackage{amsmath, amssymb}

\usepackage{xspace}
\newcommand{\model}{ARIA\xspace}
\newcommand{\vpara}[1]{\vspace{0.04in}\noindent\textbf{#1}\xspace}
\newcommand{\inputspace}{\mathcal{X}}
\newcommand{\outputspace}{\mathcal{Y}}
\newcommand{\oracle}{\mathcal{O}}
\newcommand{\modelparams}{\mathbf{\Theta}} 
\newcommand{\params}{\Theta} 

\usepackage[many]{tcolorbox} 
\usepackage{float}
\usepackage{caption} 
\usepackage{geometry}
\usepackage{afterpage}

\newcommand{\querytypes}{\mathcal{Q}}
\newcommand{\knowledgerepo}{KR}

\newcommand{\ariallm}{M_{\text{ARIA}}} 
\newcommand{\oraclellm}{M_{\text{Oracle}}} 

\usepackage{makecell}

\usepackage{amsmath,amsfonts,bm}

\def\eqref#1{equation~\ref{#1}}

\def\1{\bm{1}}

\DeclareMathAlphabet{\mathsfit}{\encodingdefault}{\sfdefault}{m}{sl}
\SetMathAlphabet{\mathsfit}{bold}{\encodingdefault}{\sfdefault}{bx}{n}

\title{Enabling Self-Improving Agents to Learn at Test Time With Human-In-The-Loop Guidance}
\author{Yufei He\textsuperscript{1}\thanks{Part of the work was done when the author was an intern at ByteDance Inc.}, Ruoyu Li\textsuperscript{2}, Alex Chen\textsuperscript{2}, Yue Liu\textsuperscript{1}, Yulin Chen\textsuperscript{1}, Yuan Sui\textsuperscript{1}, \\ \textbf{Cheng Chen\textsuperscript{2}, Yi Zhu\textsuperscript{2}, Luca Luo\textsuperscript{2}, Frank Yang\textsuperscript{2}, Bryan Hooi\textsuperscript{1}} \\
        \textsuperscript{1}National University of Singapore\\ \textsuperscript{2}ByteDance Inc.\\
        \texttt{yufei.he@u.nus.edu}}

\begin{document}
\maketitle

\begin{abstract}
Large language model (LLM) agents often struggle in environments where rules and required domain knowledge frequently change, such as regulatory compliance and user risk screening. Current approaches—like offline fine-tuning and standard prompting—are insufficient because they cannot effectively adapt to new knowledge during actual operation. To address this limitation, we propose the \textbf{A}daptive \textbf{R}eflective \textbf{I}nteractive \textbf{A}gent (ARIA)\footnote[1]{The code is available at \url{https://github.com/yf-he/aria}}, an LLM agent framework designed specifically to continuously learn updated domain knowledge at test time. ARIA assesses its own uncertainty through structured self-dialogue, proactively identifying knowledge gaps and requesting targeted explanations or corrections from human experts. It then systematically updates an internal, timestamped knowledge repository with provided human guidance, detecting and resolving conflicting or outdated knowledge through comparisons and clarification queries. We evaluate ARIA on the realistic customer due diligence name screening task on TikTok Pay, alongside publicly available dynamic knowledge tasks. Results demonstrate significant improvements in adaptability and accuracy compared to baselines using standard offline fine-tuning and existing self-improving agents. \textbf{ARIA is deployed within TikTok Pay serving over 150 million monthly active users}, confirming its practicality and effectiveness for operational use in rapidly evolving environments.
\end{abstract}
\section{Introduction}

A fundamental ability of humans is that we can learn diverse and complex skills “on the fly” (i.e., at test time), such as learning to play a new game that we have never seen before. This ability to learn “on the fly” is crucial in allowing humans to effectively perform professional tasks learned over years of experience. 

In contrast, current large language model (LLM) agents typically lack this crucial capability~\cite{bommasani2021opportunities,huang2024understanding}. Although highly effective in many scenarios thanks to large-scale pretraining and fine-tuning, existing agents are generally unable to adapt effectively once deployed~\cite{li2024personal}. 
When encountering rapidly changing domain-specific knowledge, rules, or scenarios they have never seen, these LLM-based systems frequently fail or become unreliable unless extensively retrained offline on updated labeled data~\cite{ge2023openagi}. 

An important example highlighting this challenge is customer due diligence (CDD)~\cite{mugarura2014customer} for global payment platforms—such as conducting risk list name screening~\cite{han2020artificial} for users.
An agent unable to adapt its knowledge and behavior based on these real-time changes becomes unreliable and non-compliant~\cite{bjerregaard2019danske}. 

The challenge lies in endowing agents with the capacity for continuous learning and adaptation directly during their deployment (at test time). 
To bridge this gap, 
we propose the Adaptive Reflective Interactive Agent (ARIA), a general-purpose framework designed to enable effective LLM learning at test time through structured self-assessment and human-in-the-loop interactions. Specifically, ARIA utilizes structured self-evaluation to detect its gaps or uncertainties, proactively engages human experts for targeted guidance, and systematically integrates obtained human knowledge through a carefully managed internal knowledge base organized by timestamps and flagging mechanisms for outdated or conflicting information.

ARIA is architected not just to execute tasks, but to actively manage its own knowledge limitations and collaborate with human experts. This is enabled through two core capabilities:

\vpara{Intelligent Guidance Solicitation.} 
Rather than using fixed heuristics or confidence scoring thresholds, ARIA initiates a structured internal question-and-answer self-dialogue. Upon producing an initial preliminary judgment, ARIA responds to reflective questions about the clarity and reliability of its reasoning, identifying implicit assumptions, questioning whether it possesses suitable domain knowledge, and recalling prior related experiences. 
This approach clearly highlights knowledge gaps and uncertainties, which directly motivates targeted human assistance.

\vpara{Human-Guided Knowledge Adaptation.} 
After identifying knowledge uncertainties, ARIA proactively solicits support and receives guidance—corrections, detailed explanations, or updated rules—from human domain experts. It incorporates these human-provided knowledge inputs into a structured knowledge repository that marks each knowledge item with timestamps. 
Whenever a new knowledge update occurs, ARIA retrieves related entries by semantic matching in its repository and compares them against the new information. 
If inconsistencies or contradictions between new and old knowledge appear, ARIA adjusts the status of outdated rules, clearly marking them as potentially obsolete. 
To efficiently maintain consistency, ARIA also generates active clarification queries back to human experts, resolving detected contradictions and ensuring up-to-date accuracy.

While we demonstrate ARIA's effectiveness within the context of name screening tasks, it is conceived as a general framework. 
Its core principles—enabling test-time learning through reflective uncertainty assessment and structured integration of human guidance—are applicable to a wide range of domains. 
Any task requiring strong, evolving domain-specific knowledge where human expertise is available and valuable for ongoing refinement could benefit from this approach, particularly those operating in rapidly changing environments such as legal document review, complex customer support, or scientific discovery assistance.

Our primary contributions in this work are:
\begin{itemize}[leftmargin=*,itemsep=0pt,parsep=0.2em,topsep=0.3em,partopsep=0.3em]
    \item We introduce the ARIA framework, a novel and general approach enabling agents to achieve continuous learning and adaptation at test time by leveraging human-in-the-loop guidance.
    \item We detail the core abilities underpinning ARIA: mechanisms for Intelligent Guidance Solicitation based on self-reflection and uncertainty assessment, and methods for Human-Guided Knowledge Adaptation that allow structured integration and management of human-provided knowledge over time, including conflict resolution.
    \item We validate ARIA's effectiveness through experiments on realistic CDD name screening tasks on TikTok Pay and on public datasets, demonstrating significant improvements in adaptability and reliability, and note its successful deployment in a real-world industrial setting.
\end{itemize}

\section{Related Work}


\subsection{Learning at Test Time}

Learning at test time (LTT) refers to the capacity of a machine learning model to acquire new knowledge and adapt its behavior during the inference phase, which occurs after the model has been fully trained and deployed in a real-world setting. For LLMs, common approaches include in-context learning (ICL) or few-shot learning, where the model learns from examples provided within the prompt~\cite{brown2020language, min2021metaicl, wang2023large,hou2023graphmae2,he2025unigraph,he2025unigraph2,sui2024fidelis,chen2025can,chen2025robustness}, and retrieval-augmented generation (RAG), which incorporates external knowledge retrieved based on the input~\cite{lewis2020retrieval, dong2022survey,he2024generalizing}. Other methods involve test-time fine-tuning, adjusting model parameters specifically for each incoming prompt~\cite{hubotter2024efficiently, akyurek2024surprising,sui2025meta}. In the agent context, self-learning agents aim to improve autonomously through environmental interaction~\cite{liu2025symagent,gao2025flowreasoner,chen2025mlr}.
While existing methods like ICL, RAG, test-time fine-tuning, and autonomous self-learning offer some adaptability, 
ARIA distinctively establishes a human-mediated continuous learning loop, focusing on structured knowledge integration, conflict resolution via human clarification, and persistent adaptation of an evolving knowledge base at test time.

\subsection{Human-in-the-Loop with LLMs}
Human-in-the-loop (HITL) is a collaborative and iterative approach in the field of LLM that integrates human input and expertise into various stages of the LLM system's lifecycle. 
A prominent example is reinforcement learning from human feedback (RLHF), which fine-tunes models to align their outputs with human preferences, often collected offline~\cite{rafailov2023direct, bai2022training, casper2023open,he2025evaluating}. 
Other HITL applications involve using human annotators to label data or provide feedback on model outputs to guide iterative improvements~\cite{li2025llm, yan2024practical}, or assist in specific tasks like path planning for robotic agents~\cite{xiao2023llm}.
These HITL approaches typically focus on offline alignment or use human input primarily as labels for subsequent model refinement. 
ARIA's HITL mechanism is distinct in its focus on enabling the agent to (1) intelligently initiate interaction based on self-assessed knowledge gaps, and (2) collaboratively build and maintain an evolving knowledge base through structured dialogue and feedback integration with human experts during test time. 

\begin{figure*}[t]
    \centering
    \includegraphics[width=1\linewidth]{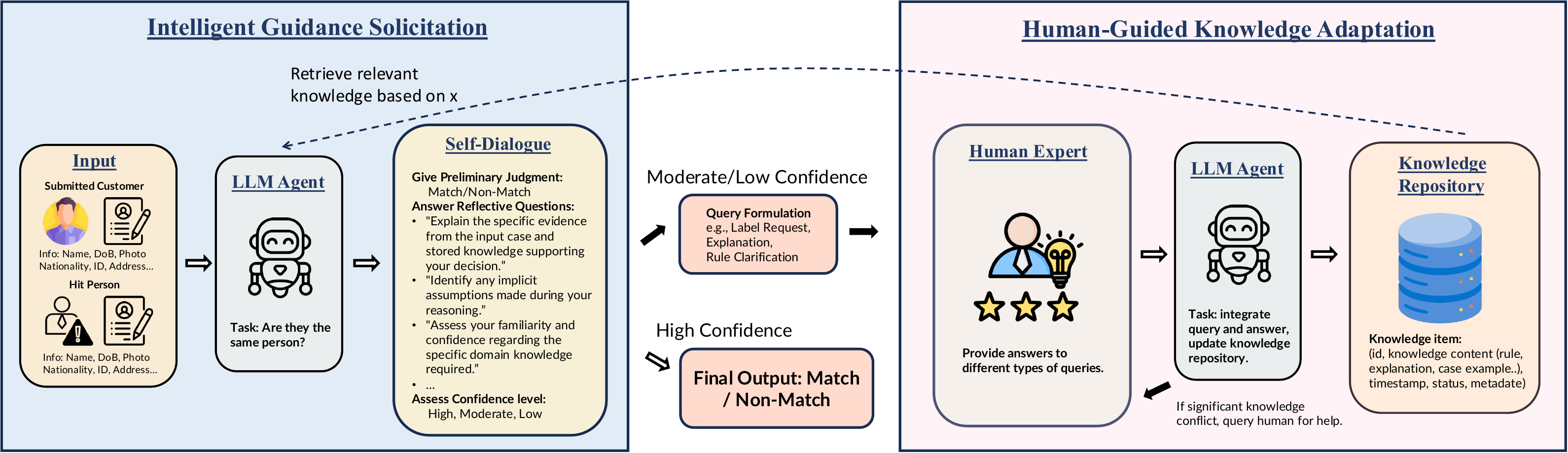}
    \caption{Overview of the ARIA framework. The agent processes input, assesses the need for guidance via self-reflection, and can solicit human expert feedback. This feedback is integrated into an evolving knowledge repository, enabling learning at test time.}
    \label{fig:arch}
    \vspace{-5mm}
\end{figure*}

\section{Problem Definition}
\label{sec:problem_setup_formal}

\subsection{Problem Statement}
The problem is to design an agent that processes a sequence of data instances $X = (x_1, x_2, \dots, x_N)$ arriving at test time. The agent must make a prediction $\hat{y}_i$ for each $x_i$. The environment may be dynamic, meaning the underlying data distribution $P(y|x)$ can change over time. The agent must adapt its internal state or model $\params$ to maintain high performance, with the ability to solicit targeted guidance from a human expert oracle $\oracle$ under a predefined interaction budget $B$.

\subsection{Formalism: Learning at Test Time with Human-in-the-Loop Guidance}
Let $\inputspace$ be the input space and $\outputspace$ be the output (label) space. The agent encounters a stream of $N$ instances $X = (x_1, x_2, \dots, x_N)$, where $x_i \in \inputspace$. The true label for $x_i$ is $y_i^* \in \outputspace$.

\vpara{Learning at Test Time (LTT):}
The agent possesses an internal state or model parameterized by $\params_i \in \modelparams$ at time step $i$ (before processing $x_i$). Its decision policy is $\pi: \inputspace \times \modelparams \rightarrow \outputspace$, producing a prediction $\hat{y}_i = \pi(x_i; \params_i)$. LTT is characterized by the update of the agent's state/model during the sequential processing of test instances:
\begin{equation}
    \params_{i+1} = f(\params_i, x_i, \hat{y}_i, q_i, h_i)
\end{equation}
where $f$ is the learning update function, $q_i$ is a query made to the human expert (if any), and $h_i$ is the feedback received from the expert. The initial state is $\params_0$. This signifies that learning occurs instance by instance as the agent operates.

\vpara{Human-in-the-Loop (HITL) Guidance:}
The agent can interact with a human expert oracle $\oracle$ to obtain guidance.
\begin{itemize}[leftmargin=*,itemsep=0pt,parsep=0.2em,topsep=0.3em,partopsep=0.3em]
    \item At each time step $i$, the agent makes a decision $d_i \in \{ \texttt{predict\_only}, \texttt{query\_expert} \}$.
    \item If $d_i = \texttt{query\_expert}$, the agent selects a query $q_i$ from a predefined set of query types $\mathcal{Q}$. 
    The set of allowed query types $\mathcal{Q}$ defines the various forms of guidance the agent can request from the human expert. Each query type $q \in \mathcal{Q}$ is designed to elicit specific information to aid the agent's learning process.
    \item Each query type $q \in \mathcal{Q}$ has an associated cost $c(q) > 0$. 
    For simplicity in experiments, $c(q)$ can be set to 1 for all $q \in \mathcal{Q}$.
    \item The total cost of queries is constrained by a budget $B$: $\sum_{j=1}^{N} c(q_j) \leq B$.
    \item If $q_i \neq \texttt{null}$, the oracle $\oracle$ provides feedback $h_i = \oracle(x_i, q_i)$. 
    This feedback $h_i$ is used in the learning update function $f$.
\end{itemize}

\vpara{Objective:}
The overall objective is to design the agent's policy $\pi$ and learning update function $f$ 
to maximize a cumulative performance metric $M_\text{perf}$ 
over the entire stream $X$, subject to the budget constraint $B$.
\begin{equation}
    \max_{\pi, f, \text{query\_strategy}} \sum_{i=1}^{N} \text{Eval}(\hat{y}_i, y_i^*) \quad \text{s.t.} \quad \sum_{j=1}^{N} c(q_j) \leq B 
\end{equation}
where $\text{Eval}(\cdot, \cdot)$ is an evaluation function.

\subsection{Instantiation in the CDD Context}
\label{sec:cdd}
This problem is instantiated in the CDD name screening task on TikTok Pay.
\begin{itemize}[leftmargin=*,itemsep=0pt,parsep=0.2em,topsep=0.3em,partopsep=0.3em]
    \item \textbf{Input Space $\inputspace$}: An agent receives pairs of user information and watchlist hit person information and determine if they refer to the same individual (\texttt{Match}) or not (\texttt{Non-Match}). A match decision typically prevents account opening. Each $x_i = (u_i, wh_i)$ is a pair of user information $u_i$ and watchlist hit information $wh_i$. These include fields such as names, aliases, native language names, nationality, address, date of birth, identification documents, and for $wh_i$, sensitive information like position or reason for listing.
    \item \textbf{Output Space $\outputspace$}: \{\texttt{Match}, \texttt{Non-Match}\}.
    \item \textbf{Data Stream}: A sequence of $N=11,846$ real-world cases, processed chronologically. The dataset is highly imbalanced, containing only 156 \texttt{Match} (Positive) cases, with the remainder being \texttt{Non-Match} (Negative).
    \item \textbf{Human Expert Oracle $\oracle$}: Real human domain experts from the TikTok's compliance teams provide responses $h_i$ (feedback). The forms of interaction include, but are not limited to, requesting case labels, explanations for decisions, resolutions for knowledge conflicts, or clarifications of rules.
    \item \textbf{Dynamic Environment}: The chronological nature of the data, coupled with the real-world source, means that underlying rules, data patterns, and watchlist characteristics may evolve, requiring the agent to adapt.
\end{itemize}

\section{Methodology}

\subsection{Overview of ARIA}
\label{sec:aria_ltt_hitl}
The agent, ARIA, processes a stream of instances $X = (x_1, x_2, \dots, x_N)$ sequentially. Its internal state, primarily a structured Knowledge Repository ($\texttt{KR}_i$), evolves from $\params_i \approx \texttt{KR}_i$ to $\params_{i+1} \approx \texttt{KR}_{i+1}$ at each time step $i$. This iterative learning process unfolds as follows:

\begin{enumerate}[leftmargin=*,itemsep=0pt,parsep=0.2em,topsep=0.3em,partopsep=0.3em]
    \item \textbf{Initial Task Processing:} For an incoming instance $x_i$, the agent, using its current knowledge repository $\texttt{KR}_i$ and its base LLM $M_\text{LLM}$, generates an initial prediction $\hat{y}_i = \pi(x_i; \texttt{KR}_i, M_\text{LLM})$ along with supporting reasoning $r_i$. The policy $\pi$ combines retrieval from $\texttt{KR}_i$ with the reasoning capabilities of $M_\text{LLM}$.
    \item \textbf{Intelligent Guidance Solicitation (IGS):} The agent performs a structured self-assessment of its preliminary judgment $(\hat{y}_i, r_i)$ and its underlying knowledge. This is denoted as $S_i = \texttt{IGS\_Assess}(\hat{y}_i, r_i, \texttt{KR}_i)$, which includes an assessed confidence level $\text{conf}_i$ and identified knowledge gaps or uncertainties $g_i$. Based on this assessment $S_i$, the agent decides $d_i \in \{\texttt{predict\_only, query\_expert}\}$. If $d_i = \texttt{query\_expert}$ and the cumulative query cost up to the previous step $\sum_{j=1}^{i-1} c(q_j) < B$ (where $B$ is the total budget), the agent then formulates a specific query $q_i = \texttt{IGS\_FormulateQuery}(S_i) \in \mathcal{Q}$. $\mathcal{Q}$ represents the set of available query types that facilitate various forms of human guidance.
    \item \textbf{Human Expert Interaction:} The query $q_i$ (if any) is presented to the human expert oracle $\oracle$, who provides feedback $h_i = \oracle(x_i, q_i)$. 
    \item \textbf{Human-Guided Knowledge Adaptation (HGKA):} The agent updates its knowledge repository from $\texttt{KR}_i$ to $\texttt{KR}_{i+1}$ using the feedback $h_i$ and the context of $x_i, \hat{y}_i, q_i$. This update $\texttt{KR}_{i+1} = \texttt{HGKA\_Update}(\texttt{KR}_i, x_i, \hat{y}_i, q_i, h_i)$ constitutes the core of ARIA's LTT ability, $f$.
\end{enumerate}
The core components, IGS and HGKA, are detailed below.

\subsection{Intelligent Guidance Solicitation}
\label{sec:igs}
The Intelligent Guidance Solicitation (IGS) module is responsible for determining when human intervention is necessary and formulating targeted queries to maximize the utility of human feedback within the budget $B$. It moves beyond simple confidence scores by enabling the agent to perform structured self-reflection. Let $t_\text{{current}}$ denote the current time or time step of processing.

\vpara{Process.}
Given an instance $x_i$, the agent's initial decision $\hat{y}_i = \pi(x_i; \texttt{KR}_i, M_\text{LLM})$, and its reasoning $r_i$, the IGS module initiates a self-reflection phase.
\begin{enumerate}[leftmargin=*,itemsep=0pt,parsep=0.2em,topsep=0.3em,partopsep=0.3em]
    \item \textbf{Structured Self-Dialogue:} The agent is prompted with a predefined set of $N_{\texttt{RQ}}$ reflective questions $\texttt{RQ} = \{\text{rq}_1, \text{rq}_2, \dots, \text{rq}_{N_{\texttt{RQ}}}\}$. These questions are designed to probe its understanding of $x_i$, the basis for $\hat{y}_i$, the assumptions made, the relevance and sufficiency of knowledge in $\texttt{KR}_i$, and consistency with past, similar instances. 
    The agent internally generates answers $\text{ans}_k = M_\text{LLM}(\text{rq}_k, x_i, \hat{y}_i, r_i, \texttt{KR}_i)$ for each $\text{rq}_k$. The collection of these question-answer pairs forms the self-dialogue $D_i^{\texttt{self}} = \{(\text{rq}_1, \text{ans}_1), \dots, (\text{rq}_{N_{\texttt{RQ}}}, \text{ans}_{N_{\texttt{RQ}}})\}$.
    \item \textbf{Confidence Self-Assessment:} Based on $D_i^{\texttt{self}}$, the agent performs a self-assessment to determine its confidence in $\hat{y}_i$. This results in an explicit confidence statement $\text{conf}_i = \texttt{AssessConfidence}(D_i^{\texttt{self}})$, where $\text{conf}_i \in \mathcal{C} = \{\texttt{High, Moderate, Low}\}$.
    \item \textbf{Intervention Trigger and Query Formulation:}
    The decision $d_i$ to query the expert is made: $d_i = \texttt{query\_expert}$ if $\text{conf}_i \in \{\texttt{Moderate, Low}\}$ and the budget constraint is not violated (i.e., $\sum_{j=1}^{i-1} c(q_j) + c(\texttt{potential } q_i) \leq B$, where $c(\texttt{potential } q_i)$ is the cost of the query to be formulated). Otherwise, $d_i = \texttt{predict\_only}$.
    If $d_i = \texttt{query\_expert}$, the agent formulates a query $q_i = \texttt{IGS\_FormulateQuery}(D_i^{\texttt{self}})$. The content of $D_i^{\texttt{self}}$ (i.e., the identified sources of uncertainty or knowledge gaps $g_i$) directly informs the type and content of $q_i$. For example, if $D_i^{\texttt{self}}$ reveals uncertainty about the correct label due to ambiguous evidence, the agent may ask for the correct label. If it identifies a lack of specific domain knowledge, it may ask for the relevant rule or an explanation. An illustrative example of this IGS process, detailing the self-dialogue and subsequent query formulation, is provided in Appendix~\ref{app:igs_example_detailed}.
\end{enumerate}
The self-dialogue $D_i^{\texttt{self}}$ is provided to the human expert alongside $q_i$, enabling them to deliver targeted and efficient guidance $h_i$. 

\subsection{Human-Guided Knowledge Adaptation}
\label{sec:hgka}
The Human-Guided Knowledge Adaptation (HGKA) module is responsible for integrating the human expert's feedback $h_i$ into the agent's knowledge repository $\texttt{KR}_i$, thereby updating its state to $\texttt{KR}_{i+1}$. This forms the learning step $\params_{i+1} = f(\params_i, \dots)$ in the LTT process, where $f$ is realized by $\texttt{HGKA\_Update}$. Let $t_\text{{current}}$ represent the current processing time or timestamp.

\vpara{Knowledge Repository Structure.}
The knowledge repository $\texttt{KR}$ is a collection of structured knowledge items $k$. Each item $k \in \texttt{KR}$ is represented as a tuple:
$k = (\text{kid}, K, ts_\text{{added}}, ts_\text{{validated}}, S, \text{M}_\text{{meta}})$
where:
\begin{itemize}[leftmargin=*,itemsep=0pt,parsep=0.2em,topsep=0.3em,partopsep=0.3em]
    \item $\text{kid}$: A unique identifier for the knowledge item.
    \item $K$: The content of the knowledge item (e.g., a rule, an explanation, a factual statement, or a case exemplar $(x_j, y_j^*, \texttt{reason}_j)$).
    \item $ts_\text{{added}}$: Timestamp of when $K$ was initially added to $\texttt{KR}$.
    \item $ts_\text{{validated}}$: Timestamp of when $K$ was last validated or updated by human feedback, or its status was changed.
    \item $S \in \{\texttt{Valid, PotentiallyOutdated, Superseded}\}$: The current validity status of $K$.
    \item $\text{M}_\text{{meta}}$: Additional metadata, such as source of $K$ (human expert, self-derived), usage frequency, links to related $\text{kid}$s (e.g., $\text{kid}_{\text{superseded\_by}}$).
\end{itemize}
The agent's state is $\params_i \approx \texttt{KR}_i$.

\vpara{Processing Feedback and Updating Knowledge.}
When human feedback $h_i$ is received for query $q_i$ concerning $x_i$, HGKA module performs $\texttt{KR}_{i+1} = \texttt{HGKA\_Update}(\texttt{KR}_i, x_i, \hat{y}_i, q_i, h_i)$ as follows:
\begin{enumerate}[leftmargin=*,itemsep=0pt,parsep=0.2em,topsep=0.3em,partopsep=0.3em]
    \item \textbf{Knowledge Item Extraction:} The feedback $h_i$ is parsed to extract a set of new explicit knowledge assertions, denoted $K_\text{{asserted}}$.
    \item \textbf{Timestamping and Initial Storage:} 
    Each extracted knowledge content $K_{\text{extracted}}$ forms a new, timestamped ($ts_{\text{added/validated}} = t_{\text{current}}$), \texttt{Valid} item with a unique $\text{kid}$ and metadata, then is provisionally added to $\texttt{KR}$.
    \item \textbf{Conflict Detection and Resolution:} For each such newly extracted knowledge content $K_\text{{extracted}}$ (which will form the content $K$ of a new knowledge item $k_\text{{new}}$):
    \begin{enumerate}[leftmargin=*,itemsep=0pt,parsep=0.2em,topsep=0.3em,partopsep=0.3em]
        \item \textbf{Retrieval of Related Knowledge:} Identify potentially related existing knowledge items $\texttt{KR}_\text{rel} = \{k_j \in \texttt{KR}_i \mid \texttt{Sim}(K_\text{{extracted}}, k_j.K) > \tau_\text{sim} \}$, where $\texttt{Sim}$ is a semantic similarity function (e.g., based on embeddings) and $\tau_\text{sim}$ is a similarity threshold.
        \item \textbf{Comparison and Status Update:} For each $k_\text{{old}} \in \texttt{KR}_\text{rel}$:
        An LLM-based comparison function $\texttt{Comp}(K_\text{{extracted}}, k_\text{{old}}.K) \rightarrow \texttt{relation}$ determines if $K_\text{{extracted}}$ contradicts, supersedes, updates, or is consistent with $k_\text{{old}}.K$.
        \begin{itemize}[leftmargin=*,itemsep=0pt,parsep=0.2em,topsep=0.3em,partopsep=0.3em]
            \item If $K_\text{{extracted}}$ supersedes $k_\text{{old}}.K$: $k_\text{{old}}.S \leftarrow \texttt{Superseded}$; $k_\text{{old}}.\text{M}_\text{{meta}}.\texttt{superseded\_by} \leftarrow k_\text{{new}}.\text{kid}$; $k_\text{{old}}.ts_\text{{validated}} \leftarrow t_\text{{current}}$.
            \item If $K_\text{{extracted}}$ conflicts with $k_\text{{old}}.K$ making $k_\text{{old}}.K$ uncertain: $k_\text{{old}}.S \leftarrow \texttt{PotentiallyOutdated}$; $k_\text{{old}}.ts_\text{{validated}} \leftarrow t_\text{{current}}$.
        \end{itemize}
        The updated $k_\text{{old}}$ items and the new knowledge item $k_\text{{new}}$ (containing $K_\text{{extracted}}$) become part of $\texttt{KR}_{i+1}$. An example of this conflict detection and resolution process is provided in Appendix~\ref{app:conflict_example_detailed}.
    \end{enumerate}
    \item \textbf{Active Clarification Query Generation:} If comparison reveals unresolvable ambiguity/conflict (e.g., contradictory expert advice), HGKA generates an internal clarification query $q'_{\text{new}}$. IGS may then issue $q'_{\text{new}}$ to the human expert $\oracle$, subject to budget $B$. (Example: Appendix~\ref{app:clarify_example_detailed})
\end{enumerate}
The resulting updated collection of knowledge items forms $\texttt{KR}_{i+1}$.

\vpara{Temporally-Informed Knowledge Retrieval.} To process an instance $x_j$ at current time $t_{\text{current}}$, the agent's policy $\pi(x_j; \texttt{KR}_j, M_{\text{LLM}})$ utilizes a relevant knowledge subset $\texttt{KR}_{\text{subset}}$. This subset is retrieved by scoring and ranking items $k \in \texttt{KR}_j$ based on three factors: a validity weight $W_S(k.S)$ (where $W_S(\texttt{Valid})=1.0$, $W_S(\texttt{PotentiallyOutdated})=w_{\text{po}}$, and $W_S(\texttt{Superseded})=0.0$), a recency score $S_T(k, t_{\text{current}}) = \exp(-\lambda \cdot (t_{\text{current}} - k.ts_{\text{validated}}))$ reflecting the timeliness of $k.ts_{\text{validated}}$, and a semantic relevance score $S_R(k, x_j)$ quantifying the contextual pertinence of $k.K$ to $x_j$. These are combined into a composite score:
\begin{equation}
\begin{aligned}
    Score(k, x_j, t_{\text{current}}) = W_S(k.S) \times \\ S_T(k, t_{\text{current}}) \times S_R(k, x_j)
\end{aligned}
\end{equation}
This multiplicative approach penalizes outdated, old, or irrelevant items. $\texttt{KR}_{\text{subset}}$ is then formed by selecting either the top-$N_{\text{top}}$ items or all items exceeding a score threshold $\tau_{\text{score}}$, ensuring decisions are guided by the most current, valid, and contextually appropriate knowledge.

\begin{table*}[t]\small
  \centering
  \caption{\textbf{Overall performance comparison on the TikTok Pay for varying query budgets ($B$).} The dataset consists of a chronological sequence of $N=11,846$ real-world cases, highly imbalanced with only 156 \texttt{Match} (Positive) cases and 11,690 \texttt{Non-Match} (Negative) cases. Best results for interactive methods at each budget are highlighted.}
  \label{tab:overall_performance_budget_matched}
  \vspace{-2mm}
  \begin{tabular}{ll  cccc  cccc }
    \toprule[1.1pt]
    \multirow{2}{*}{\textbf{Method}} & \multirow{2}{*}{\textbf{Model}} & \multicolumn{4}{c}{\textbf{Sensitivity}} & \multicolumn{4}{c}{\textbf{Specificity}} \\
    \cmidrule(lr){3-6} \cmidrule(lr){7-10} 
     & & B=50 & B=100 & B=500 & B=1000 & B=50 & B=100 & B=500  & B=1000  \\
    \midrule
    Static Agent & Qwen2.5-7B  & \multicolumn{4}{c}{0.6474} & \multicolumn{4}{c}{0.6124}  \\
    Static Agent & GPT-4o  & \multicolumn{4}{c}{0.7051} & \multicolumn{4}{c}{0.6539}  \\
    Offline Fine-tuning & Qwen2.5-7B & 0.6410 & 0.6603 & 0.6731 & 0.6987 & 0.6317 & 0.6492 & 0.6776 & 0.6791 \\
    RAG Agent & GPT-4o & 0.7756 & 0.8013 & 0.8141 & 0.8333 & 0.6864 & 0.7051 & 0.7308 & 0.7462 \\
    \midrule
    \multicolumn{10}{l}{\textbf{\textit{Self-Improving Agents}}} \\
    Self-Refine & GPT-4o  & \multicolumn{4}{c}{0.7244} & \multicolumn{4}{c}{0.6821}  \\
    Reflexion & GPT-4o  & \multicolumn{4}{c}{0.7692} & \multicolumn{4}{c}{0.6902}  \\
    Multi-Agent Debate & GPT-4o  & \multicolumn{4}{c}{0.7628} & \multicolumn{4}{c}{0.6970}  \\
    \midrule
    \multicolumn{10}{l}{\textbf{\textit{Active Learning Methods}}} \\
    Random Querying & GPT-4o & 0.7372 & 0.7949 & 0.8205 & 0.8590 & 0.6725 & 0.6994 & 0.7410 & 0.7667 \\
    Simple Uncertainty & GPT-4o & 0.7884 & 0.8013 & 0.8590 & 0.8718 & 0.6936 & 0.7218 & 0.7590 & 0.7853 \\
    \midrule
    ARIA (ours) & Qwen2.5-7B & 0.7564 & 0.7756 & 0.8077 & 0.8397 & 0.6859 & 0.7154 & 0.7549 & 0.7795 \\
    ARIA (ours) & GPT-4o & \textbf{0.8013} & \textbf{0.8333} & \textbf{0.8653} & \textbf{0.8910} & \textbf{0.7151} & \textbf{0.7423} & \textbf{0.7810} & \textbf{0.8026} \\
    \bottomrule[1.1pt]
  \end{tabular}
  \vspace{-4mm}
\end{table*}

\begin{table}[t]\small
\centering
\renewcommand\tabcolsep{4.9pt}
\caption{\textbf{Ablation studies on ARIA key components.}}
\vskip -0.1in
\label{tab:kc}
\begin{tabular}{lcc}
\toprule[1.1pt]
  Method (B=100)  & Sensitivity & Specificity  \\
\midrule
    \model & \textbf{0.8333} & \textbf{0.7423}\\
    Labels-Only ARIA & 0.7949 & 0.7139 \\
    w/o Self-Dialogue &  0.8141 & 0.7319    \\
    w/o KR Conflict Resolution &  0.8012 & 0.7128    \\
    w/o Temporally-Informed KR &  \textbf{0.8333} & 0.7341     \\
\bottomrule[1.1pt]
\end{tabular}
\vspace{-3.3mm}
\end{table}

\begin{table}[t]\small
\centering
\renewcommand\tabcolsep{4.6pt}
\caption{\textbf{Efficiency comparison of the ARIA model and Human Experts.}}
\vskip -0.1in
\label{tab:effici}
\begin{tabular}{lccc}
\toprule[1.1pt]
    Method & Sensitivity & Specificity & AHT \\
\midrule
    Human Experts & 1.0 & 1.0 & 12min \\
    \model (B=50) & 0.8013 & 0.7151 & 0.13min \\
    \model (B=100) & 0.8333 & 0.7423 & 0.15min \\
    \model (B=500) & 0.8653 & 0.7810 & 0.20min \\
    \model (B=1000) & 0.8910 & 0.8026 & 0.23min \\
    \multirow{2}{*}{\makecell[l]{\model w/ Full Oracle \\ Access (B=3121)}} &  \multirow{2}{*}{0.9428} &  \multirow{2}{*}{0.8814} & \multirow{2}{*}{0.41min} \\ 
   & & & \\
\bottomrule[1.1pt]
\end{tabular}
\vspace{-3.3mm}
\end{table}

\begin{table*}[t]\small
  \centering
  \caption{\textbf{Overall performance on the CUAD dataset for clause type identification.} The dataset consists of a stream of $N = 13,101$ contract clauses across 41 types. Best results for interactive methods at each budget are highlighted.}
  \vspace{-3mm}
  \label{tab:cuad_performance}
  \begin{tabular}{llcccccc}
    \toprule[1.1pt]
    \multirow{2}{*}{\textbf{Method}} & \multirow{2}{*}{\textbf{Model}} & \multicolumn{5}{c}{\textbf{Accuracy}} \\
    \cmidrule(lr){3-7} 
     & & B=50 & B=100 & B=500 & B=1000 & B=2000 &   \\
    \midrule
    Static Agent  & Qwen2.5-7B  & \multicolumn{5}{c}{0.3515}   \\ 
    Static Agent  & GPT-4o  & \multicolumn{5}{c}{0.4872} \\ %
    \midrule
    Offline Fine-tuning & Qwen2.5-7B & 0.3680 & 0.3918 & 0.4317 & 0.4721 & 0.4909 \\ 
    RAG Agent & GPT-4o & 0.4953 & 0.5101 & 0.5309 & 0.5597 & 0.5735 \\ 
    \midrule
    \multicolumn{7}{l}{\textbf{\textit{Self-Improving Agents}}} \\
    Self-Refine & GPT-4o  & \multicolumn{5}{c}{0.4931}  \\
    Reflexion & GPT-4o  & \multicolumn{5}{c}{0.4995}   \\
    Multi-Agent Debate & GPT-4o  & \multicolumn{5}{c}{0.4890}  \\
    \midrule
    \multicolumn{7}{l}{\textbf{\textit{Active Learning Methods}}} \\
    Random Querying & GPT-4o & 0.4901 & 0.4983 & 0.5154 & 0.5338 & 0.5492 \\ 
    Simple Uncertainty & GPT-4o & 0.4975 & 0.5116 & 0.5353 & 0.5604 & 0.5789 \\ 
    \midrule
    ARIA (ours) & Qwen2.5-7B & 0.3801 & 0.4196 & 0.4703 & 0.5117 & 0.5435 \\
    ARIA (ours) & GPT-4o & \textbf{0.5084} & \textbf{0.5397} & \textbf{0.5781} & \textbf{0.6072} & \textbf{0.6358} \\ %
    \bottomrule[1.1pt]
  \end{tabular}
  \vspace{-4mm}
\end{table*}

\section{Deployment on TikTok Pay}
We evaluate ARIA on CDD name screening task on TikTok Pay, as introduced in Section~\ref{sec:cdd} and Appendix~\ref{sec:cddappe}.

\subsection{Baselines}
ARIA interacts with a human expert oracle $\mathcal{O}$ up to a budget $B$. 
For fair comparison, Offline Fine-tuning and RAG baselines are prepared before deployment using knowledge equivalent to budget $B$. Active learning baselines also interact with $\mathcal{O}$ up to budget $B$ but employ different query strategies.

\vpara{Static Agent}: An LLM agent with general knowledge, using a fixed initial policy $\pi_0$ and no task-specific adaptation.

\vpara{Offline Fine-tuning}: An agent fine-tuned once before deployment on data equivalent to budget $B$. 

\vpara{RAG Agent}: An LLM agent using a static knowledge base (KB) populated before deployment with data equivalent to budget $B$. 

\vpara{Active Learning (Random Querying)}: Queries the human expert oracle $\mathcal{O}$ by selecting cases randomly up to budget $B$.

\vpara{Active Learning (Simple Uncertainty Sampling)}: Queries $\mathcal{O}$ up to budget $B$ based on a standard uncertainty sampling heuristic (e.g., low confidence).

\vpara{Self-Refine}~\cite{madaan2023self}: An LLM iteratively refining its own output by generating an initial response, providing self-feedback, and then improving the response based on that feedback.

\vpara{Reflexion}~\cite{shinn2023reflexion}:
Agent improves itself by verbally reflecting on past task feedback. These reflections are stored in memory to guide subsequent decision-making.

\vpara{Multi-Agent Debate}~\cite{du2023improving}: This approach uses multiple LLM agents that learn from each others' feedback to collaboratively refine solutions through iterative debate. 

For fair comparison, we use GPT-4o~\cite{hurst2024gpt} and Qwen2.5-7B~\cite{qwen2.5} as base LLMs for all baselines and our method.
\subsection{Evaluation Metrics}
We evaluate performance using:

\vpara{Sensitivity:} The proportion of actual \texttt{Match} cases that are correctly identified as \texttt{Match}.

\vpara{Specificity:} The proportion of actual \texttt{Non-Match} cases that are correctly identified as \texttt{Non-Match}.

\subsection{Results}
The performance comparison on the TikTok Pay application (Table~\ref{tab:overall_performance_budget_matched}) reveals several key insights.
1) ARIA, particularly with GPT-4o, consistently outperforms other methods across all query budgets ($B$) in both Sensitivity and Specificity, showcasing its superior adaptability through effective human-in-the-loop guidance.
2) ARIA demonstrates more significant performance gains with increasing query budgets compared to other active learning strategies, indicating more efficient use of human expertise. For instance, ARIA (GPT-4o) at $B=1000$ achieves 0.8910 Sensitivity and 0.8026 Specificity, notably surpassing Simple Uncertainty (0.8718 Sensitivity, 0.7853 Specificity).
3) ARIA effectively enhances the performance of both stronger (GPT-4o) and weaker (Qwen2.5-7B) base models, often outperforming static or self-improving agents reliant on GPT-4o. This underscores ARIA's test-time learning abilities and its advantage in integrating real-time human feedback. Some case examples can be found in Appendix~\ref{sec:examples}.

\subsection{Model Analysis}
\vpara{Ablation on Key Components.}
Table~\ref{tab:kc} presents an ablation study on ARIA's key components. Restricting ARIA to `\text{Labels-Only}' queries, thereby foregoing other types of human guidance, diminished its effectiveness and underscored the value of comprehensive feedback. 
The model operating `\text{w/o Self-Dialogue}', thus lacking the agent's structured self-reflection for uncertainty assessment, showed a clear reduction in performance. 
Similarly, the absence of the KR Conflict Resolution mechanism in the `\text{w/o KR Conflict Resolution}' variant, vital for maintaining a coherent knowledge base, resulted in a substantial performance drop. 
Operating `\text{w/o Temporally-Informed KR}', which omits the prioritization of recent and relevant knowledge, also impacted ARIA's precision. Collectively, these results affirm the critical role of each evaluated component in ARIA's overall success.

\vpara{Efficiency Analysis.}
As shown in Table~\ref{tab:effici}, the \model\ demonstrates significant efficiency gains compared to conventional human expert evaluation in CDD name screening tasks. The previous online method for this business was solely human evaluation. In real business, human experts typically take around 12 minutes for Average Handling Time (AHT) per case. 
In contrast, \model's AHT is substantially lower, even as the query budget (B) increases. 
Even when \model\ is allowed to query as many questions as possible (B=3121, referred to as Full Oracle Access), its AHT is only 0.41 minutes. 
This indicates a considerable time saving. 
It's also noted that for human experts, reviewing a case from scratch is time-consuming, whereas answering a query from an agent is much faster.

\section{Experiments on Public Dataset}
\subsection{Setup}
We evaluate ARIA in the domain of legal text analysis using the publicly available Contract Understanding Atticus Dataset (CUAD)~\cite{hendrycks1cuad}, which includes over 500 commercial contracts annotated with 41 clause types. ARIA sequentially processes extracted clauses to identify their types and assess potential risks.

To enable Learning at Test Time (LTT) with Human-in-the-Loop (HITL), we simulate the expert oracle ($\mathcal{O}$) using a powerful LLM, providing scalable human-like feedback. Clauses are streamed chronologically, with simulated concept drifts introduced via shifting clause distributions and evolving oracle responses.

ARIA's performance is assessed on clause identification accuracy, adaptability to dynamic changes, and budget efficiency. We compare against static baselines and ARIA variants with limited oracle access. Detailed settings, including data preprocessing, oracle prompting, dynamic simulation, metrics, and baselines, are provided in Appendix~\ref{app:detailed_setup}.

\subsection{Results}

Experiments on the CUAD dataset, presented in Table~\ref{tab:cuad_performance}, further highlight ARIA's efficacy in test-time learning.
1) The necessity of LTT with HITL for complex, multi-class legal understanding is evident: ARIA (GPT-4o) achieves a remarkable 0.6358 accuracy at $B=2000$, a substantial leap from the 0.4872 accuracy of the static GPT-4o agent, demonstrating its capability to adapt where pre-trained knowledge falls significantly short.
2) ARIA's structured human interaction and dynamic knowledge management prove superior to autonomous adaptation or static retrieval strategies. ARIA (GPT-4o) consistently outperforms self-improving agents (e.g., Reflexion at 0.4995) and the RAG agent (0.5735 at $B=2000$ equivalent pre-population), underscoring the value of its targeted guidance solicitation in navigating the nuances of evolving legal interpretations and clause variations.
3) The framework demonstrates efficient knowledge acquisition and scalability. Even with a modest budget ($B=500$), ARIA (GPT-4o) reaches 0.5781 accuracy, surpassing the RAG agent with a much larger implicit budget. Furthermore, ARIA significantly elevates the performance of the weaker Qwen2.5-7B model (0.5435 accuracy at $B=2000$), making it competitive with several GPT-4o based methods, which validates the robustness of ARIA's learning mechanisms.

\section{Conclusion}

This paper introduces ARIA, an LLM agent framework for test-time learning through human-in-the-loop guidance. ARIA addresses conventional model limitations in dynamic environments by assessing uncertainty via self-dialogue, soliciting expert corrections, and updating a timestamped, conflict-resolving knowledge base. Experiments on the name screening task in TikTok Pay and with public datasets demonstrate significant improvements in adaptability. ARIA's principles are broadly applicable to domains requiring evolving knowledge and human expertise, paving the way for more robust and reliable AI agents.


\section*{Limitations}

While ARIA demonstrates promising results in enabling agents to learn at test time with human-in-the-loop guidance, several limitations warrant discussion.

First, the effectiveness of ARIA is intrinsically linked to the \textbf{availability, quality, and scalability of human expertise}. The framework assumes access to responsive and accurate human experts. In scenarios with very high query volumes, or where expert feedback is delayed, inconsistent, or erroneous, ARIA's learning capability and overall performance could be significantly impacted. The practical cost and logistical challenges of maintaining a pool of readily available experts for diverse and evolving tasks are also important considerations not fully explored~\cite{sui2024can}.

Second, the \textbf{complexity of knowledge representation and conflict resolution} could pose challenges as the knowledge repository (KR) grows in size and intricacy. While ARIA incorporates mechanisms for timestamping and managing conflicting information, highly nuanced, subtly contradictory, or deeply contextual expert guidance might be difficult to integrate perfectly. Ensuring the long-term coherence and accuracy of a large, evolving KR, and preventing the accumulation of outdated or overly specific knowledge, remains an ongoing research area.

Third, regarding \textbf{generalizability}, ARIA has been primarily validated on tasks like customer due diligence and legal text analysis. These domains, while dynamic, often involve relatively structured information and specific types of uncertainty. The framework's adaptability and the efficacy of its current self-reflection and knowledge adaptation mechanisms in vastly different domains—such as those requiring complex common-sense reasoning, creative generation, or interaction with the physical world—would require further investigation and potentially significant modifications to the query types and self-dialogue structures~\cite{liu2025efficient,wang2025safety}.

Fourth, the evaluation on the public CUAD dataset relied on an \textbf{LLM-simulated human expert oracle}. Although this approach facilitates scalable experimentation, it may not fully replicate the nuances, potential biases, occasional errors, or the depth of insight that a genuine human domain expert would provide. The dynamics of interaction and the nature of guidance from a simulated oracle might differ from real-world human-agent collaboration, potentially affecting the observed learning patterns.

Finally, the \textbf{efficiency of the self-dialogue and knowledge management processes} could become a concern in applications with extremely high throughput or stringent real-time constraints. While crucial for ARIA's adaptability, the computational overhead associated with structured self-reflection, semantic retrieval from the KR, and conflict resolution mechanisms might need further optimization for certain deployment scenarios. The current study focuses more on the effectiveness of learning rather than a detailed analysis of computational performance under heavy load.

\section*{Ethical Considerations}
A key ethical consideration revolves around the human experts involved in ARIA's learning loop. In business contexts, these individuals are paid, well-trained employees. While ARIA is designed to augment their capabilities and improve efficiency, the increasing proficiency of such AI systems raises concerns about the long-term impact on their roles. 
There is a potential for over-reliance on the automated system, which could lead to a deskilling of these trained employees over time if their direct engagement with complex decision-making diminishes. 
Furthermore, as ARIA demonstrates significant efficiency gains, there is an inherent risk that such technology could be perceived or utilized as a means to reduce the human workforce, leading to job displacement for these skilled professionals. Therefore, careful consideration must be given to deploying ARIA in a manner that genuinely collaborates with and empowers human experts, focusing on handling increased complexity or volume, rather than solely as a replacement strategy. This includes fostering new skills, redefining job roles to work alongside AI, and ensuring that the benefits of automation are shared equitably.

\bibliography{ref}

\appendix

\tableofcontents
\addcontentsline{toc}{section}{Appendix}

\section{Details of Deployment on TikTok Pay}
\label{sec:cddappe}
\subsection{Task Background}
We use the task of Customer Due Diligence (CDD) name screening for TikTok Pay as a running example. In this scenario, the agent assists human experts by evaluating new customer applications against various risk factors, primarily focusing on risk list screening. This domain faces frequent updates to regulations and watchlists, inherent ambiguity in data (e.g., name variations), and requires nuanced interpretation, making continuous learning essential.
The typical workflow involves:
\begin{enumerate}[leftmargin=*,itemsep=0pt,parsep=0.2em,topsep=0.3em,partopsep=0.3em]
    \item A user submits personal information (name, date of birth, address, etc.) for account opening.
    \item A retrieval system queries large databases (e.g., risk lists) and returns potential matches ("hits") based on the submitted information.
    \item An agent receives pairs of user information and hit information and must determine if they refer to the same individual (Match) or not (Non-match). A match decision typically prevents account opening.
\end{enumerate}
This task is challenging due to incomplete or inconsistent information in both user submissions and database entries, as well as ambiguous or frequently changing screening rules (e.g., due to regulatory updates). 
Simply providing all policies and rules within a large prompt context to an LLM is impractical due to the inherent ambiguity in complex regulatory texts which LLMs may struggle to interpret and apply consistently, especially when rules conflict. 

Consequently, the current industry convention often relies heavily on manual auditing of most cases by human experts. While ensuring accuracy, this approach consumes significant time and financial resources. These limitations underscore the need for a more adaptive and collaborative approach like ARIA, which seeks to leverage automation while intelligently engaging human expertise where it is most needed.

\subsection{Reflective Questions}
Below is a list of example reflective questions ($rq_k \in RQ$) an agent might use for self-assessment. These questions are designed to probe the agent's understanding of the input case ($x_i$), the basis for its preliminary judgment ($\hat{y}_i$), any implicit assumptions made, the relevance and sufficiency of its stored knowledge ($KR_i$), and consistency with past, similar instances.

\begin{itemize}[leftmargin=*,itemsep=0pt,parsep=0.2em,topsep=0.3em,partopsep=0.3em]
    \item Explain the specific evidence from the input case and stored knowledge supporting your decision. 
    \item Identify any implicit assumptions made during your reasoning.
    \item Assess your familiarity and confidence regarding the specific domain knowledge required (e.g., 'How familiar am I with company policy on acceptable DOB discrepancies? Do I know the rules for matching Chinese name variations?').
    \item Compare this case to similar past experiences and assess the consistency of your reasoning. 
    \item Based on the input case $x_i = \text{``...''}$, my preliminary judgment is $\hat{y}_{\text{type}}$. What is my confidence level (High/Moderate/Low) for this judgment, and why? 
    \item Which specific phrases or keywords in the input case $x_i$ support this classification? Are there any conflicting indicators within the case? 
    \item After retrieving relevant items from my knowledge repository $KR_i$, how consistent is my preliminary judgment $\hat{y}_{\text{type}}$ with these items (e.g., definitions, exemplars, rules)? 
    \item What are the key obligations and permissions implied by the input case $x_i$ if it is indeed classified as $\hat{y}_{\text{type}}$?
    \item Is my knowledge regarding the predicted type $\hat{y}_{\text{type}}$ (including definitions and rules) in my knowledge repository $KR_i$ marked as recently validated, or is it potentially outdated? 
\end{itemize}

\subsection{Baselines}
To evaluate the specific contributions of ARIA's components, we compare its performance against several baseline models. ARIA interacts with the human expert oracle $\mathcal{O}$ up to budget $B$ during its run. For fair comparison, the Offline Fine-tuning and RAG baselines are provided \textbf{before deployment} with knowledge derived from an equivalent set of human interactions (representing the same budget $B$). The active learning baselines interact during their run, similar to ARIA, but use different query strategies.

\vpara{Static Agent (No Prior Exposure):} An LLM agent initialized with general knowledge. It processes all cases $x_i$ using its fixed initial policy $\pi_0$. 

\vpara{Offline Fine-tuning (Pre-Deployment):} This agent is fine-tuned \textit{once before deployment} on the labeled examples and explanations derived from the human interaction set (equivalent to budget $B$). After deployment, it operates as a \textbf{static model}, using the policy learned during this single pre-training phase. 

\vpara{RAG Agent (Static Populated KB):} An LLM agent employing Retrieval-Augmented Generation (RAG). Its static knowledge base is populated \textit{before deployment} with the rules, explanations, and labeled examples derived from the same set of human interactions (equivalent to budget $B$) available to ARIA and the Fine-tuned agent. During the test run, it retrieves from this fixed knowledge base to generate decisions but \textbf{cannot update} the KB or resolve conflicts dynamically. 

\vpara{Active Learning (Random Querying):} This agent operates similarly to ARIA by querying the human expert oracle $\mathcal{O}$ during the test run, up to the budget $B$. However, it selects cases $x_i$ to query \textbf{randomly}, without using any intelligent strategy based on uncertainty or self-reflection. It uses the feedback (e.g., labels) to update its internal state (e.g., for few-shot prompting). 

\vpara{Active Learning (Simple Uncertainty Sampling):} Like the random querying agent, this baseline interacts with the expert oracle $\mathcal{O}$ during the run up to budget $B$. It decides when to query based on a standard \textbf{uncertainty sampling heuristic} (e.g., querying when the prediction confidence score is below a threshold $\theta$). This compares ARIA's structured self-reflection against simpler, common active learning query strategies for utilizing the budget $B$.

\vpara{Self-Refine}~\cite{madaan2023self}: This approach enables a language model to iteratively improve its own outputs without requiring additional training data or separate models. The core idea involves the model generating an initial response, then critically evaluating that response to provide feedback to itself, and subsequently using this feedback to generate a refined output. This feedback-refinement loop can be repeated to enhance the quality of the final response.

\vpara{Reflexion}~\cite{shinn2023reflexion}: This framework allows language agents to learn from past experiences through verbal reinforcement rather than by updating their underlying model weights. Reflexion agents reflect on feedback received from tasks (which can be simple scores or textual critiques), generate textual self-reflections, and store these in an episodic memory. This memory of past reflections then helps guide the agent to make better decisions and improve its performance in subsequent attempts.

\vpara{Multi-Agent Debate}~\cite{du2023improving}: This method utilizes multiple language model instances, or "agents," to collaboratively solve a problem or arrive at an answer. The agents individually generate initial responses and then engage in a structured debate over one or more rounds. During the debate, agents can present their reasoning, critique the outputs of other agents, and refine their own positions based on the collective discussion. This process aims to improve the accuracy and robustness of the final outcome by leveraging diverse perspectives and encouraging critical evaluation.

\section{Illustrative Examples of ARIA's Mechanisms}
\label{app:aria_examples_detailed}

\begin{figure*}[t] 
\centering
\begin{tcolorbox}[colframe=brown!100!black, colback=brown!10, title=\textbf{Example: Intelligent Guidance Solicitation in Action}, width=\textwidth]
\textbf{Task:} Compare User (Li Xiaoming, DOB 1985-03-12) with Watchlist Hit (Li Xiao Ming, DOB 1985-03-12, Nationality Unknown).

\textbf{Agent Initial Decision:} "True Positive (Match). The names are identical when ignoring spacing, and the DOB matches exactly."

\textbf{Self-Reflection Dialogue (Agent's internal process)}:
\begin{itemize}
    \item Q: Explain evidence. A: Name `Li Xiaoming` vs `Li Xiao Ming` is a direct match considering common spacing variations based on a known rule. DOB `1985-03-12` matches exactly.
    \item Q: Identify assumptions. A: Assumed the rule about ignoring spacing is universally applicable here. Assumed lack of nationality information on the hit is not a disqualifier for a match if name/DOB are strong, as I don't have an explicit rule for this scenario.
    \item Q: Assess domain knowledge familiarity. A: I am familiar with the spacing rule. However, I lack an explicit rule or clear knowledge about how missing nationality impacts match confidence according to current policy. My confidence in handling this aspect is moderate.
    \item Q: Compare to past cases. A: I have seen similar name/DOB matches. No directly comparable past cases with missing nationality and a high-confidence True Positive outcome were found in my recent experiences.
\end{itemize}
\textbf{Agent Self-Assessed Confidence:} "[B] Moderate confidence with specific uncertainties. While name and DOB match strongly, I lack certainty regarding the current policy on handling missing nationality information in watchlist hits."

\textbf{Outcome:} The case is flagged for human review. The human expert receives the initial decision, reasoning, and the agent's stated uncertainty ("lack certainty regarding the current policy on handling missing nationality"). The agent effectively asks for the correct label and clarification on the rule for missing nationality. The expert can then provide the correct judgment and clarify the specific rule.
\end{tcolorbox}
\captionof{figure}{Illustrative example of the Intelligent Guidance Solicitation (IGS) process.}
\label{fig:igs_example_appendix} 
\end{figure*}

\begin{figure*}[t]
\centering
\begin{tcolorbox}[colframe=brown!100!black, colback=brown!10, title=\textbf{Example: Conflict Detection and Resolution}, width=\textwidth]
\textbf{Scenario:} The agent's knowledge base contains an old rule provided weeks ago. A human expert now provides an updated rule.
\begin{itemize}
    \item \textbf{Existing Knowledge ($k_{old}$):} \texttt{\{ID: Rule\_045, Timestamp: 2025-04-10, Content: "Exact pinyin match is required for Chinese names.", Status: Valid\}}
    \item \textbf{New Human Feedback ($K_{new}$ from $h_i$):} \texttt{\{ID: Rule\_123, Timestamp: 2025-05-05, Content: "For Chinese names, minor pinyin variations (e.g., 'Zhang' vs 'Zang') are acceptable if other identifiers (like DOB) match closely. Exact match is no longer strictly required.", Status: Valid\}}
\end{itemize}
\textbf{Process:}
\begin{enumerate}
    \item \textbf{Retrieval:} Semantic search for the content of the new feedback retrieves the old rule (Rule\_045) due to topic overlap ("pinyin", "Chinese names", "match").
    \item \textbf{Comparison:} The LLM compares the new and old rules. It identifies that the new rule explicitly allows variations, directly contradicting the old rule's requirement for an exact match. The new rule states the previous rule is no longer required.
    \item \textbf{Status Update:} The system updates the status of the old rule: \\
    \texttt{\{ID: Rule\_045, Timestamp: 2025-04-10, Content: "Exact pinyin match is required for Chinese names.", Status: Superseded by Rule\_123 on 2025-05-05\}}
\end{enumerate}
\textbf{Outcome:} The old rule (Rule\_045) is marked as superseded and linked to the new rule (Rule\_123). The knowledge base now reflects the latest guidance.
\end{tcolorbox}
\captionof{figure}{Illustrative example of the Conflict Detection and Resolution process within HGKA.}
\label{fig:hgka_conflict_example_appendix} 
\end{figure*}

\begin{figure*}[t]
\centering
\begin{tcolorbox}[colframe=brown!100!black, colback=brown!10, title=\textbf{Example: Active Clarification Query Generation}, width=\textwidth]
\textbf{Scenario:} The knowledge base contains an old rule. An expert provides a new, somewhat ambiguous rule.
\begin{itemize}
    \item \textbf{Existing Knowledge ($k_{old}$):} \texttt{\{ID: Rule\_045, ..., Content: "Exact pinyin match required for Chinese names.", ..., Status: Valid\}}
    \item \textbf{New Human Feedback ($K_{new}$ from $h_i$):} \texttt{\{ID: Rule\_124, Timestamp: 2025-05-05, Content: "Allow minor pinyin variations for common names like 'Zhang/Zang' if DOB is exact.", Status: Valid\}}
\end{itemize}
\textbf{HGKA Process - Comparison Step:}
The LLM compares the new rule with the old one. It detects a conflict but notes the new rule seems conditional ("common names," "exact DOB"). It's unclear if this completely replaces the old rule or is a specific exception.

\textbf{Active Clarification Query Generation:}
The system formulates a question for the human expert:
"Human Expert, you provided a new rule (Rule\_124) allowing pinyin variations for common names with exact DOBs. Does this rule \textit{only} apply under these specific conditions, or does it relax the general requirement for exact pinyin matches (from the old Rule\_045) more broadly? Should the old Rule\_045 now be considered outdated in all cases, or only when the conditions of the new Rule\_124 are met?"

\textbf{Outcome:} This question is queued for the next human interaction (or raised immediately if appropriate). The expert's answer will allow the system to precisely update the status of the old rule (e.g., mark it as 'Superseded' or 'Valid but modified by Rule\_124') and ensure the knowledge base is accurate.
\end{tcolorbox}
\captionof{figure}{Illustrative example of the Active Clarification query generation process within HGKA.}
\label{fig:hgka_clarify_example_appendix} 
\end{figure*}

\subsection{Example: Intelligent Guidance Solicitation (IGS) in Action}
\label{app:igs_example_detailed} 
The following example~\ref{fig:igs_example_appendix} illustrates the IGS process as described in Section~\ref{sec:igs}.

\subsection{Example: Conflict Detection and Resolution}
\label{app:conflict_example_detailed} 
The example~\ref{fig:hgka_conflict_example_appendix} demonstrates the conflict detection and resolution mechanism within HGKA, as described in Section~\ref{sec:hgka}.

\subsection{Example: Active Clarification Query Generation}
\label{app:clarify_example_detailed} 
The example~\ref{fig:hgka_clarify_example_appendix} illustrates how HGKA can generate a query for active clarification, as discussed in Section~\ref{sec:hgka}.

\section{ARIA CDD Task Case Examples}
\label{sec:examples}

\textbf{Please note: All personal information data and review rules in the examples presented in this appendix and the main text (including all data details) are fictional or have been desensitized for illustrative purposes only and do not represent real user data or complete actual rules.}

\subsection{Case 1: Malay Name with Patronymic and DOB Discrepancy}
The following example~\ref{fig:aria_cdd_malay_name_example1}~\ref{fig:aria_cdd_malay_name_example2}~\ref{fig:aria_cdd_malay_name_example3} illustrates the whole ARIA process.

\begin{figure*}[p] 
\centering
\begin{tcolorbox}[colframe=brown!75!black, colback=brown!10, title=\textbf{Example: CDD Case Review - Malay Name and DOB}, width=\textwidth, fonttitle=\bfseries]

\textbf{1. Input Data:}
\begin{itemize}
    \item \textbf{User Application:}
    \begin{itemize}
        \item \textbf{Name:} Siti Aishah binti Hamid
        \item \textbf{Date of Birth (DOB):} 12/05/1985
        \item \textbf{Nationality:} Malaysian
    \end{itemize}
    \item \textbf{Watchlist Hit:}
    \begin{itemize}
        \item \textbf{Name:} Aishah Hamid
        \item \textbf{Alias:} Siti Hamid
        \item \textbf{DOB:} May 1985
        \item \textbf{Nationality:} Malaysian
        \item \textbf{Reason for Listing:} Politically Exposed Person (PEP) - Level 3
    \end{itemize}
\end{itemize}

\vspace{2mm}
\textbf{2. ARIA's Review Process:}

\textbf{A. Initial Task Processing (LLM Agent):}
\begin{itemize}
    \item \textbf{Preliminary Judgment:} Potential Match.
    \item \textbf{Reasoning:} "The core name components 'Aishah' and 'Hamid' appear in both the user application and the watchlist hit. The user's name 'Siti Aishah binti Hamid' includes 'Siti' (common female first name or title) and 'binti Hamid' (patronymic indicating 'daughter of Hamid'). The watchlist name 'Aishah Hamid' could be a shortened version. The DOBs show a match for month and year (May 1985), with the day missing from the watchlist. Nationality matches."
\end{itemize}

\end{tcolorbox}
\captionof{figure}{Part1: Illustrative example of ARIA's review process for a CDD case involving Malay name structure and DOB discrepancy.}
\label{fig:aria_cdd_malay_name_example1}
\end{figure*}

\begin{figure*}[p] 
\centering
\begin{tcolorbox}[colframe=brown!75!black, colback=brown!10, title=\textbf{Example: CDD Case Review - Malay Name and DOB}, width=\textwidth, fonttitle=\bfseries]

\textbf{B. Intelligent Guidance Solicitation (IGS):}
\begin{itemize}
    \item \textbf{Self-Dialogue (Agent's internal process, based on Reflective Questions):}
    \begin{itemize}
        \item \textbf{Q: Explain evidence supporting decision.}
        \item \textbf{A:} "Core names 'Aishah', 'Hamid' match. DOB Month/Year 'May 1985' matches. Nationality 'Malaysian' matches. Watchlist alias 'Siti Hamid' also shares components."
        \item \textbf{Q: Identify assumptions made.}
        \item \textbf{A:} "Assumed 'Siti' can be a prefix/title and 'binti Hamid' can be dropped or is represented by 'Hamid' as a surname in the watchlist context. Assumed DOB day missing from watchlist is acceptable if Month/Year match for a PEP Level 3."
        \item \textbf{Q: Assess domain knowledge familiarity for Malay names and DOB policies.}
        \item \textbf{A:} "Familiar with general name matching. Current knowledge base ($KR_i$) has a rule: `Rule\_078: Malay patronymics (bin/binti) may be absent in simplified name versions. Core given names and father's name (as surname) are key. Timestamp: 2024-11-10, Status: Valid`. However, confidence in handling 'Siti' specifically as a title versus part of a given name in conjunction with PEP Level 3 and partial DOB is moderate. Current policy on acceptable DOB discrepancy for PEP Level 3 is not explicitly clear for day-level mismatch when month/year match."
        \item \textbf{Q: Compare to similar past cases.}
        \item \textbf{A:} "Have seen cases of partial DOB matches resolved as True Match for lower risk profiles. No direct high-confidence precedent for PEP Level 3 with this specific combination of Malay name structure and day-level DOB ambiguity."
    \end{itemize}
    \item \textbf{Confidence Self-Assessment:} Moderate. "While core elements align, the combination of Malay name structure ('Siti', 'binti'), partial DOB, and PEP Level 3 status introduces uncertainty regarding strictness of current matching policy."
    \item \textbf{Decision:} Query Human Expert.
    \item \textbf{Query Formulation to Human Expert:} "Potential Match: User 'Siti Aishah binti Hamid' (DOB 12/05/1985) vs Watchlist 'Aishah Hamid' (DOB May 1985, PEP L3). Core names and DOB Month/Year match. My knowledge 'Rule\_078' supports patronymic variations. \textbf{Questions:} 1. Is the match valid considering 'Siti' and missing DOB day for a PEP L3? 2. Please confirm/clarify current policy for DOB day discrepancy for PEP L3 when Month/Year match and Malay name structure is involved."
\end{itemize}

\textbf{C. Human Expert Interaction:}
\begin{itemize}
    \item \textbf{Human Expert Feedback ($h_i$):} "1. Yes, this is a True Match. 'Siti' is a common honorific/first name component and 'Aishah binti Hamid' is often shortened to 'Aishah Hamid'. 2. For PEP Level 3, a Month/Year DOB match is sufficient if other identifiers (like name components and nationality) strongly align. The day difference is acceptable in this context. This is policy revision `Pol\_Update\_DOB\_PEP3\_v2\_20250515`."
\end{itemize}

\end{tcolorbox}
\captionof{figure}{Part2: Illustrative example of ARIA's review process for a CDD case involving Malay name structure and DOB discrepancy.}
\label{fig:aria_cdd_malay_name_example2}
\end{figure*}

\begin{figure*}[p] 
\centering
\begin{tcolorbox}[colframe=brown!75!black, colback=brown!10, title=\textbf{Example: CDD Case Review - Malay Name and DOB}, width=\textwidth, fonttitle=\bfseries]

\textbf{C. Human Expert Interaction:}
\begin{itemize}
    \item \textbf{Human Expert Feedback ($h_i$):} "1. Yes, this is a True Match. 'Siti' is a common honorific/first name component and 'Aishah binti Hamid' is often shortened to 'Aishah Hamid'. 2. For PEP Level 3, a Month/Year DOB match is sufficient if other identifiers (like name components and nationality) strongly align. The day difference is acceptable in this context. This is policy revision `Pol\_Update\_DOB\_PEP3\_v2\_20250515`."
\end{itemize}

\textbf{D. Human-Guided Knowledge Adaptation (HGKA):}
\begin{itemize}
    \item \textbf{Knowledge Item Extraction from $h_i$:}
    \begin{itemize}
        \item $K_{asserted1}$: "Case (User: Siti Aishah binti Hamid, DOB 12/05/1985; WL: Aishah Hamid, DOB May 1985, PEP L3) is a True Match." (as an exemplar)
        \item $K_{asserted2}$: "Policy `Pol\_Update\_DOB\_PEP3\_v2\_20250515`: For PEP Level 3, Month/Year DOB match is sufficient if other strong identifiers align; day difference is acceptable."
    \end{itemize}
    \item \textbf{Timestamping and Initial Storage:} New items $k_{new1}$ (for $K_{asserted1}$) and $k_{new2}$ (for $K_{asserted2}$) added to KR with $ts_{added/validated}$ = current date/time, Status: Valid.
    \item \textbf{Conflict Detection and Resolution:}
    \begin{itemize}
        \item Agent searches KR for rules related to "DOB PEP Level 3". Finds existing (hypothetical) rule: `Rule\_102: PEP Level 3 requires exact DOB match. Timestamp: 2024-01-20, Status: Valid`.
        \item Comparison: $K_{asserted2}$ (new policy) contradicts `Rule\_102`.
        \item Status Update: `Rule\_102.S` $\leftarrow$ Superseded; `Rule\_102.M\_{meta}.superseded\_by` $\leftarrow k_{new2}$; `Rule\_102.ts\_{validated}` $\leftarrow$ current date/time.
    \end{itemize}
    \item \textbf{Final Knowledge Repository ($KR_{i+1}$):} Contains the new exemplar $k_{new1}$, the new validated policy $k_{new2}$, and the updated (superseded) `Rule\_102`. Existing `Rule\_078` on Malay names remains Valid and reinforced by the exemplar.
\end{itemize}

\vspace{2mm}
\textbf{3. Final Output (after expert guidance for this case):} True Match.

\end{tcolorbox}
\captionof{figure}{Part3: Illustrative example of ARIA's review process for a CDD case involving Malay name structure and DOB discrepancy.}
\label{fig:aria_cdd_malay_name_example3}
\end{figure*}

\subsection{Case 2: Name Transliteration and Fuzzy DOB (Year Only)}
The following example, illustrated across Figure~\ref{fig:aria_cdd_translit_example1}, Figure~\ref{fig:aria_cdd_translit_example2}, and Figure~\ref{fig:aria_cdd_translit_example3}, demonstrates ARIA's process for a case involving name transliteration and a year-only DOB match against a risks list.

\subsection{Case 3: Name with Initials, DOB Transposition, and Address Correlation}
The following example, illustrated across Figure~\ref{fig:aria_cdd_initials_example1}, Figure~\ref{fig:aria_cdd_initials_example2}, and Figure~\ref{fig:aria_cdd_initials_example3}, demonstrates ARIA's process for a case involving a name with initials, a potential DOB transposition, and address correlation against a financial fraud watchlist.

\clearpage
\begin{figure*}[p] 
\centering
\begin{tcolorbox}[colframe=blue!75!black, colback=blue!10, title=\textbf{Example: CDD Case Review - Name Transliteration \& Year-Only DOB}, width=\textwidth, fonttitle=\bfseries]

\textbf{1. Input Data:}
\begin{itemize}
    \item \textbf{User Application:}
    \begin{itemize}
        \item \textbf{Name:} Khalid Sheikh Al-Maktoum
        \item \textbf{Date of Birth (DOB):} 22/07/1965
        \item \textbf{Nationality:} Qatari
    \end{itemize}
    \item \textbf{Watchlist Hit (risks List):}
    \begin{itemize}
        \item \textbf{Name:} Al Maktum, Khalid S.
        \item \textbf{Alias:} Khaled Al Maktoom
        \item \textbf{DOB:} 1965
        \item \textbf{Nationality:} (Not Specified)
        \item \textbf{Reason for Listing:} International risks (Terrorism Financing)
    \end{itemize}
\end{itemize}

\vspace{2mm}
\textbf{2. ARIA's Review Process:}

\textbf{A. Initial Task Processing (LLM Agent):}
\begin{itemize}
    \item \textbf{Preliminary Judgment:} Potential High-Risk Match.
    \item \textbf{Reasoning:} "The names show similarity: 'Khalid' matches, 'Al-Maktoum' appears as 'Al Maktum' and 'Al Maktoom' (common transliteration variations). The watchlist name 'Al Maktum, Khalid S.' suggests a Westernized name order (Surname, Given Name Initial). The 'Sheikh' component is a common honorific and might be omitted in formal records. The DOB year '1965' matches. The watchlist reason is a high-risk risks listing."
\end{itemize}

\end{tcolorbox}
\captionof{figure}{Part 1: Input Data and Initial Task Processing for Name Transliteration Case.}
\label{fig:aria_cdd_translit_example1}
\end{figure*}

\clearpage
\begin{figure*}[p] 
\centering
\begin{tcolorbox}[colframe=blue!75!black, colback=blue!10, title=\textbf{Example: CDD Case Review - Name Transliteration \& Year-Only DOB (Continued)}, width=\textwidth, fonttitle=\bfseries]

\textbf{B. Intelligent Guidance Solicitation (IGS):}
\begin{itemize}
    \item \textbf{Self-Dialogue (Agent's internal process):}
    \begin{itemize}
        \item \textbf{Q: Explain evidence supporting decision.}
        \item \textbf{A:} "Primary name 'Khalid' and 'Al-Maktoum' (with variations 'Al Maktum', 'Al Maktoom') align. DOB year '1965' matches. Alias 'Khaled Al Maktoom' reinforces name similarity despite spelling variation."
        \item \textbf{Q: Identify assumptions made.}
        \item \textbf{A:} "Assumed 'Al Maktum, Khalid S.' is an inverted name order. Assumed 'Sheikh' is an honorific and its absence is acceptable. Assumed transliteration differences ('Al-Maktoum' vs. 'Al Maktum'/'Al Maktoom', 'Khalid' vs 'Khaled') are within acceptable variance for this type of name. Assumed year-only DOB match is potentially sufficient for a high-risk risks hit."
        \item \textbf{Q: Assess domain knowledge familiarity for Arabic name transliteration and risks list DOB policies.}
        \item \textbf{A:} "My knowledge base ($KR_i$) contains: `Rule\_042: Arabic names often have 'Al' prefix which can be written with or without hyphen, or separated. Timestamp: 2024-10-01, Status: Valid`. Also, `Rule\_065: Transliteration of Arabic names can result in multiple Latin script spellings. Timestamp: 2024-09-15, Status: Valid`. However, the specific tolerance for combined transliteration variance AND inverted order AND year-only DOB for a high-risk terrorism financing risk is not explicitly covered by a high-confidence rule. Current policy on DOB precision for such high-risk alerts is critical."
        \item \textbf{Q: Compare to similar past cases.}
        \item \textbf{A:} "Processed cases with transliteration issues and year-only DOB matches but typically for lower-risk alerts or with more secondary identifiers. This combination for a terrorism financing risk is novel or has low precedent in recent experience."
    \end{itemize}
    \item \textbf{Confidence Self-Assessment:} Moderate to Low (leaning Low due to high risk). "Significant name component overlap and DOB year match. However, the combination of name order inversion, multiple transliteration points, honorific omission, and year-only DOB for a very high-risk watchlist entry requires careful validation against precise current policy."
    \item \textbf{Decision:} Query Human Expert.
    \item \textbf{Query Formulation to Human Expert:} "Potential High-Risk Match: User 'Khalid Sheikh Al-Maktoum' (DOB 22/07/1965) vs Watchlist 'Al Maktum, Khalid S.' (DOB 1965, Terrorism Financing risk). Name elements 'Khalid', 'Al-Maktoum' (as Al Maktum/Al Maktoom) and DOB year match. Assumed transliteration, name inversion, and honorific omission. \textbf{Questions:} 1. Are these variations collectively acceptable for a True Match given the high-risk nature? 2. What is the current policy for DOB precision (year-only vs. full DOB) for terrorism financing risks when primary name elements show strong alignment but also variations?"
\end{itemize}

\end{tcolorbox}
\captionof{figure}{Part 2: Intelligent Guidance Solicitation for Name Transliteration Case.}
\label{fig:aria_cdd_translit_example2}
\end{figure*}

\clearpage
\begin{figure*}[p] 
\centering
\begin{tcolorbox}[colframe=blue!75!black, colback=blue!10, title=\textbf{Example: CDD Case Review - Name Transliteration \& Year-Only DOB (Continued)}, width=\textwidth, fonttitle=\bfseries]

\textbf{C. Human Expert Interaction:}
\begin{itemize}
    \item \textbf{Human Expert Feedback ($h_i$):} "1. Yes, this is a confirmed True Match. The name variations (Khalid/Khaled, Al-Maktoum/Al Maktum/Al Maktoom, omission of 'Sheikh', inverted order 'Al Maktum, Khalid S.') are common and well within tolerance for this individual based on our enhanced due diligence records for this specific entity. 2. For high-priority risks like terrorism financing, if strong name correlation exists across multiple reliable sources, a year-only DOB match from the risks list is acceptable as an initial hit, triggering immediate escalation and further due diligence, which has confirmed this link. Refer to `Policy\_SAN\_DOB\_HighRisk\_v3\_20250410`."
\end{itemize}

\textbf{D. Human-Guided Knowledge Adaptation (HGKA):}
\begin{itemize}
    \item \textbf{Knowledge Item Extraction from $h_i$:}
    \begin{itemize}
        \item $K_{asserted1}$: "Exemplar: User 'Khalid Sheikh Al-Maktoum' (DOB 22/07/1965) is a True Match to Watchlist 'Al Maktum, Khalid S.' (DOB 1965, Terrorism Financing), considering specified name variations and honorific omission."
        \item $K_{asserted2}$: "Rule/Policy `Policy\_SAN\_DOB\_HighRisk\_v3\_20250410`: For high-priority risks (e.g., terrorism financing), a year-only DOB match is acceptable if strong name correlation (considering transliteration, order, honorifics based on reliable intelligence) is present; triggers escalation."
    \end{itemize}
    \item \textbf{Timestamping and Initial Storage:} New items $k_{new1}$ (exemplar) and $k_{new2}$ (policy) added to KR. Status: Valid.
    \item \textbf{Conflict Detection and Resolution:}
    \begin{itemize}
        \item Agent searches KR for rules on "risks DOB precision". Finds (hypothetical) `Rule\_S88: All risks list matches require at least Day-Month-Year or Day-Year match if name is an exact phonetic match. Timestamp: 2023-11-05, Status: Valid`.
        \item Comparison: $k_{new2}$ (new policy allowing year-only for high-risk with strong name match) provides a more specific and updated context than `Rule\_S88`.
        \item Status Update: `Rule\_S88.S` $\leftarrow$ PotentiallyOutdated (or Superseded if the new rule is explicitly stated as a replacement for such cases). Agent might flag for review or set `Rule\_S88` as superseded for "high-priority risks" context, while it might remain valid for other risk types. For this example, let's assume it's marked PotentiallyOutdated and linked to $k_{new2}$ for context. `Rule\_S88.ts\_validated` updated.
    \end{itemize}
    \item \textbf{Final Knowledge Repository ($KR_{i+1}$):} Contains $k_{new1}$, $k_{new2}$. `Rule\_042` and `Rule\_065` on Arabic names and transliteration are reinforced. `Rule\_S88` is updated.
\end{itemize}

\vspace{2mm}
\textbf{3. Final Output (after expert guidance for this case):} True Match (High-Risk, Escalate).

\end{tcolorbox}
\captionof{figure}{Part 3: Human Expert Interaction, Knowledge Adaptation, and Final Output for Name Transliteration Case.}
\label{fig:aria_cdd_translit_example3}
\end{figure*}

\begin{figure*}[p] 
\centering
\begin{tcolorbox}[colframe=green!65!black, colback=green!10, title=\textbf{Example: CDD Case Review - Initials, DOB Transposition, Address}, width=\textwidth, fonttitle=\bfseries]

\textbf{1. Input Data:}
\begin{itemize}
    \item \textbf{User Application:}
    \begin{itemize}
        \item \textbf{Name:} Michael Anthony Johnson
        \item \textbf{Date of Birth (DOB):} 02/11/1982 (DD/MM/YYYY)
        \item \textbf{Address:} Apt 4B, 789 River Road, Metropolis, USA
        \item \textbf{Nationality:} American
    \end{itemize}
    \item \textbf{Watchlist Hit (Financial Fraud List):}
    \begin{itemize}
        \item \textbf{Name:} Johnson, M. A.
        \item \textbf{Alias:} Mike Johnson
        \item \textbf{DOB:} 11/02/1982 (Potentially MM/DD/YYYY or transposed)
        \item \textbf{Address:} 780-799 River Rd, Metropolis
        \item \textbf{Nationality:} USA
        \item \textbf{Reason for Listing:} Indicted for Securities Fraud
    \end{itemize}
\end{itemize}

\vspace{2mm}
\textbf{2. ARIA's Review Process:}

\textbf{A. Initial Task Processing (LLM Agent):}
\begin{itemize}
    \item \textbf{Preliminary Judgment:} Strong Potential Match.
    \item \textbf{Reasoning:} "The surname 'Johnson' matches. The watchlist name 'Johnson, M. A.' uses initials 'M. A.' which correspond to 'Michael Anthony'. The alias 'Mike Johnson' is a common nickname for Michael Johnson. The DOB '02/11/1982' (user) and '11/02/1982' (watchlist) share the same day, month, and year components, suggesting a possible DD/MM vs MM/DD transposition. The addresses ('Apt 4B, 789 River Road, Metropolis' vs '780-799 River Rd, Metropolis') are highly similar and geographically consistent. Nationality matches. The reason for listing is serious (Securities Fraud)."
\end{itemize}

\end{tcolorbox}
\captionof{figure}{Part 1: Input Data and Initial Task Processing for Name with Initials \& DOB Transposition Case.}
\label{fig:aria_cdd_initials_example1}
\end{figure*}

\clearpage
\begin{figure*}[p] 
\centering
\begin{tcolorbox}[colframe=green!65!black, colback=green!10, title=\textbf{Example: CDD Case Review - Initials, DOB Transposition, Address (Continued)}, width=\textwidth, fonttitle=\bfseries]

\textbf{B. Intelligent Guidance Solicitation (IGS):}
\begin{itemize}
    \item \textbf{Self-Dialogue (Agent's internal process):}
    \begin{itemize}
        \item \textbf{Q: Explain evidence supporting decision.}
        \item \textbf{A:} "Surname 'Johnson' exact match. Initials 'M. A.' match 'Michael Anthony'. Nickname 'Mike' consistent. DOB components (02, 11, 1982) are identical but potentially transposed. Address 'River Road, Metropolis' matches with close street number range. Nationality matches."
        \item \textbf{Q: Identify assumptions made.}
        \item \textbf{A:} "Assumed 'M. A. Johnson' refers to 'Michael Anthony Johnson'. Assumed the DOB difference is a DD/MM/YYYY vs MM/DD/YYYY transposition. Assumed address '789 River Road' falls within '780-799 River Rd' and is a valid correlation."
        \item \textbf{Q: Assess domain knowledge familiarity for initial matching, DOB transposition, and address fuzziness.}
        \item \textbf{A:} "My knowledge base ($KR_i$) includes: `Rule\_015: Surnames with matching initials for given names are a strong indicator. Timestamp: 2024-08-20, Status: Valid`. Also, `Rule\_028: Common nicknames (e.g., Mike for Michael) should be considered. Timestamp: 2024-07-10, Status: Valid`. For DOB, `Rule\_D05: Investigate potential DD/MM vs MM/DD transposition if day component is <=12 and month component is <=12 and other identifiers are strong. Timestamp: 2024-09-01, Status: Valid`. For addresses, `Rule\_A12: Address range matches (e.g., 100-150 Main St for 125 Main St) are considered partial matches. Timestamp: 2024-06-05, Status: Valid`. The combination of all these partial/fuzzy indicators for a financial fraud watchlist needs careful assessment."
        \item \textbf{Q: Compare to similar past cases.}
        \item \textbf{A:} "Have encountered cases with DOB transpositions and address ranges that were confirmed matches. The strength here comes from multiple factors aligning (name initials, nickname, DOB elements, address proximity, nationality)."
    \end{itemize}
    \item \textbf{Confidence Self-Assessment:} Moderate to High. "Multiple strong indicators align, each supported by existing knowledge items. The key uncertainty is whether the combined strength of these slightly imperfect matches is sufficient for a definitive True Match without further clarification on the DOB transposition interpretation under current policy."
    \item \textbf{Decision:} Query Human Expert for confirmation, particularly on DOB interpretation policy.
    \item \textbf{Query Formulation to Human Expert:} "Strong Potential Match: User 'Michael Anthony Johnson' (DOB 02/11/1982, Addr: Apt 4B, 789 River Road) vs Watchlist 'Johnson, M. A.' (DOB 11/02/1982, Addr: 780-799 River Rd, Securities Fraud). Name (initials, nickname), DOB elements (potentially transposed DD/MM vs MM/DD), address, and nationality align. My rules `Rule\_015`, `Rule\_028`, `Rule\_D05`, `Rule\_A12` support these partial matches. \textbf{Questions:} 1. Is the DOB transposition (02/11 vs 11/02 for 1982) acceptable as a True Match in this context? 2. Considering all factors, should this be confirmed as True Match?"
\end{itemize}

\end{tcolorbox}
\captionof{figure}{Part 2: Intelligent Guidance Solicitation for Name with Initials \& DOB Transposition Case.}
\label{fig:aria_cdd_initials_example2}
\end{figure*}

\clearpage
\begin{figure*}[p] 
\centering
\begin{tcolorbox}[colframe=green!65!black, colback=green!10, title=\textbf{Example: CDD Case Review - Initials, DOB Transposition, Address (Continued)}, width=\textwidth, fonttitle=\bfseries]

\textbf{C. Human Expert Interaction:}
\begin{itemize}
    \item \textbf{Human Expert Feedback ($h_i$):} "1. Yes, the DOB transposition (02/11/1982 vs 11/02/1982) is a known issue with some data entry systems and is acceptable when other primary identifiers like name and address show strong correlation, especially since both day and month values are $\leq 12$. 2. This is a confirmed True Match. The combination of name (surname, initials, common nickname), transposed DOB, highly correlated address, and matching nationality meets the criteria for this watchlist. Policy `Pol\_DOB\_Transposition\_v1\_20250315` allows this under strong corroborating evidence."
\end{itemize}

\textbf{D. Human-Guided Knowledge Adaptation (HGKA):}
\begin{itemize}
    \item \textbf{Knowledge Item Extraction from $h_i$:}
    \begin{itemize}
        \item $K_{asserted1}$: "Exemplar: User 'Michael Anthony Johnson' (DOB 02/11/1982, Addr: Apt 4B, 789 River Road) is a True Match to Watchlist 'Johnson, M. A.' (DOB 11/02/1982, Addr: 780-799 River Rd, Securities Fraud), given name initials, nickname, DOB transposition, and address correlation."
        \item $K_{asserted2}$: "Policy `Pol\_DOB\_Transposition\_v1\_20250315`: DOB day/month transposition (e.g., DD/MM/YYYY vs MM/DD/YYYY) is acceptable if day \& month components are $\leq 12$ and other primary identifiers (name, address) strongly corroborate the match."
    \end{itemize}
    \item \textbf{Timestamping and Initial Storage:} New items $k_{new1}$ (exemplar) and $k_{new2}$ (policy) added to KR. Status: Valid.
    \item \textbf{Conflict Detection and Resolution:}
    \begin{itemize}
        \item Agent searches KR for rules on "DOB errors". Existing `Rule\_D05` already suggests investigating transposition but doesn't explicitly state its acceptability as a sole criterion for a True Match without ambiguity.
        \item Comparison: $k_{new2}$ (new policy) provides a more definitive and actionable guideline, formalizing the acceptability of transposition under specific conditions. It doesn't directly conflict with `Rule\_D05` but rather refines and strengthens it.
        \item Status Update: `Rule\_D05` might be updated to reference $k_{new2}$ for definitive conditions or its description enhanced. $k_{new2}$ becomes the primary reference for this specific scenario.
    \end{itemize}
    \item \textbf{Final Knowledge Repository ($KR_{i+1}$):} Contains $k_{new1}$, $k_{new2}$. Existing rules (`Rule\_015`, `Rule\_028`, `Rule\_D05`, `Rule\_A12`) are reinforced and potentially annotated with link to the new explicit policy $k_{new2}$.
\end{itemize}

\vspace{2mm}
\textbf{3. Final Output (after expert guidance for this case):} True Match.

\end{tcolorbox}
\captionof{figure}{Part 3: Human Expert Interaction, Knowledge Adaptation, and Final Output for Name with Initials \& DOB Transposition Case.}
\label{fig:aria_cdd_initials_example3}
\end{figure*}

\section{Experiment Setup for Public Dataset}
\label{app:detailed_setup}
\subsection{Dataset: CUAD (Contract Understanding Atticus Dataset)}
\label{sec:dataset}

\subsubsection{Description and Suitability}
For this experimental evaluation, we will utilize the \textbf{Contract Understanding Atticus Dataset (CUAD)} v1~\cite{hendrycks1cuad}.
\begin{itemize}[leftmargin=*,itemsep=0pt,parsep=0.2em,topsep=0.3em,partopsep=0.3em]
    \item \textbf{Source:} Available publicly via the Atticus AI Project (\url{https://www.atticusprojectai.org/cuad}).
    \item \textbf{Content:} CUAD comprises 510 full commercial legal contracts, which have been meticulously annotated by legal professionals. These annotations highlight specific segments of text corresponding to 41 distinct categories of important legal clauses (e.g., "Indemnity," "Confidentiality," "Governing Law," "Termination," "Force Majeure"). In total, the dataset contains over 13,000 annotations.
\end{itemize}

\subsection{Preprocessing and Stream Generation}
\label{ssec:stream_generation}
The data stream $X = (x_1, x_2, \dots, x_N)$ for ARIA will be constructed as follows:
\begin{enumerate}[leftmargin=*,itemsep=0pt,parsep=0.2em,topsep=0.3em,partopsep=0.3em]
    \item \textbf{Instance Definition ($x_i$):} Each instance $x_i \in \inputspace$ will be the textual content of a single contract clause. We will iterate through each of the 510 contracts. For each contract, we extract the text spans corresponding to the CUAD annotations for the 41 clause categories. Each such extracted text span constitutes an instance $x_i$.
    \item \textbf{Primary True Label ($y_i^*$):} The primary true label for an instance $x_i$ is its CUAD-annotated clause category (e.g., "Indemnity"). This will be denoted $y_{i, \text{type}}^*$.
    \item \textbf{Stream Order:}
        \begin{itemize}[leftmargin=*,itemsep=0pt,parsep=0.2em,topsep=0.3em,partopsep=0.3em]
            \item Contracts will be processed in a fixed (e.g., alphabetical by filename) order.
            \item Within each contract, clauses will be processed in the order they appear in the document.
            \item This creates a reproducible, chronologically processed stream of $N$ clause instances.
        \end{itemize}
    \item \textbf{Total Instances ($N$):} The total number of instances will be the sum of all identified clauses from all contracts, expected to be in the range of 10,000-13,000.
\end{enumerate}

\subsection{Instantiation in the LTT with HITL Guidance Framework}
\label{sec:instantiation}
We now formally map the legal clause analysis task using CUAD to the "Learning at Test Time with Human-in-the-Loop Guidance" problem statement (Section~\ref{sec:problem_setup_formal} of the ARIA description).

\begin{itemize}[leftmargin=*,itemsep=0pt,parsep=0.2em,topsep=0.3em,partopsep=0.3em]
    \item \textbf{Input Space ($\inputspace$):} The space of all possible legal clause texts. Each $x_i$ is a string representing a clause.
    \item \textbf{Output Space ($\outputspace$):} The predicted clause type $\hat{y}_{\text{type}}$ from the 41 CUAD categories.
    \item \textbf{Data Stream ($X$):} As defined in Section~\ref{ssec:stream_generation}.
    \item \textbf{True Labels ($y_i^*$):} Primarily $y_{i, \text{type}}^*$ (the CUAD clause category).
    \item \textbf{Human Expert Oracle ($\oracle$):} Simulated by a powerful LLM, $\oraclellm$ (details in Section~\ref{sec:llm_oracle}).
    \item \textbf{Query Types ($\querytypes$) and Costs ($c(q)$):} The specific queries ARIA can make to the $\oraclellm$ are:
    \begin{enumerate}[label=Q\arabic*., wide, labelindent=0pt]
        \item $q_{\text{type\_label}}(x_i)$: Request true clause type for $x_i$.
            \begin{itemize}[leftmargin=*,itemsep=0pt,parsep=0.2em,topsep=0.3em,partopsep=0.3em]
                \item Oracle Response ($h_i$): The ground truth $y_{i, \text{type}}^*$ from CUAD.
                \item Cost $c(q_{\text{type\_label}}) = 1$ unit.
            \end{itemize}
        \item $q_{\text{explain\_type}}(x_i, y_{i, \text{type}}^*)$: Request explanation for why $x_i$ belongs to $y_{i, \text{type}}^*$.
            \begin{itemize}[leftmargin=*,itemsep=0pt,parsep=0.2em,topsep=0.3em,partopsep=0.3em]
                \item Oracle Response ($h_i$): Textual explanation citing keywords, legal concepts, and contextual cues from $x_i$.
                \item Cost $c(q_{\text{explain\_type}}) = 2$ units.
            \end{itemize}
        \item $q_{\text{summarize\_rules}}(x_i, y_{i, \text{type}}^*)$: Request a summary of obligations, permissions, or key provisions as structured rules.
            \begin{itemize}[leftmargin=*,itemsep=0pt,parsep=0.2em,topsep=0.3em,partopsep=0.3em]
                \item Oracle Response ($h_i$): A list of concise rules, e.g., "\texttt{IF [condition] THEN Party A MUST [action]}".
                \item Cost $c(q_{\text{summarize\_rules}}) = 3$ units.
            \end{itemize}
        \item $q_{\text{clarify\_conflict}}(k_{\text{old}}, k_{\text{new}}, x_i)$: Request clarification if newly proposed knowledge $k_{\text{new}}$ (e.g., from oracle feedback or ARIA's own derivation) conflicts with existing knowledge $k_{\text{old}} \in KR_i$ relevant to $x_i$.
            \begin{itemize}[leftmargin=*,itemsep=0pt,parsep=0.2em,topsep=0.3em,partopsep=0.3em]
                \item Oracle Response ($h_i$): Explanation resolving the conflict, possibly by invalidating $k_{\text{old}}$, modifying $k_{\text{new}}$, or providing a contextual rule.
                \item Cost $c(q_{\text{clarify\_conflict}}) = 2$ units.
            \end{itemize}
    \end{enumerate}
    \item \textbf{Interaction Budget ($B$):} A predefined total query cost allowed over the entire stream $X$. Experiments will vary $B$ (e.g., $0.05 \sum c(q_{\text{type\_label}})$, $0.1 \sum c(q_{\text{type\_label}})$, etc., effectively a percentage of "free label" queries, scaled by average query costs if other query types are used).
    \item \textbf{Learning Update Function ($f$):} This is embodied by ARIA's Human-Guided Knowledge Adaptation (HGKA) module, which updates $\knowledgerepo_i$ to $\knowledgerepo_{i+1}$ based on $x_i, \hat{y}_i, q_i, h_i$.
    \item \textbf{Performance Metric ($M_{\text{perf}}$) and Evaluation ($\text{Eval}(\cdot, \cdot)$):}
        \begin{itemize}[leftmargin=*,itemsep=0pt,parsep=0.2em,topsep=0.3em,partopsep=0.3em]
            \item The primary performance metric will be the cumulative Accuracy for clause type identification: $M_{\text{perf}} = \sum_{i=1}^{N} \text{Accuracy}(\hat{y}_{i, \text{type}}, y_{i, \text{type}}^*)$.
        \end{itemize}
    \item \textbf{Dynamic Environment Simulation:}
        To assess ARIA's adaptability, the stream will be divided into $K$ phases. Concept drift will be simulated by:
        \begin{enumerate}[leftmargin=*,itemsep=0pt,parsep=0.2em,topsep=0.3em,partopsep=0.3em]
            \item \textit{Changing Clause Frequencies:} The probability distribution $P(y_{\text{type}})$ of encountering different clause types will be altered between phases. For example, an early phase might be rich in "Confidentiality" clauses, while a later phase might see a surge in "Data Security" or "Force Majeure" clauses.
            \item \textit{Evolving Clause Phrasing (Subtle Textual Drift):} In later phases, for a subset of clause types, instances $x_i$ can be subtly rephrased (e.g., using another LLM as a paraphraser) while preserving legal meaning. This tests ARIA's robustness to linguistic variations.
            \item \textit{Changing Oracle Interpretations (Simulated Policy Drift):} The $\oraclellm$'s prompting for $q_{\text{summarize\_rules}}$ for specific clause types can be modified between phases. For example:
                \begin{quote}
                Phase 1 Oracle on rule for "Governing Law": "Standard interpretation: Choice of law is absolute."
                Phase 2 Oracle on rule for "Governing Law": "Recent precedent *Case Z* suggests public policy exceptions are more broadly applied to choice of law provisions."
                \end{quote}
                This directly tests ARIA's HGKA module in updating its $\knowledgerepo$ based on evolving expert guidance.
            \item \textit{Introduction of Novel (Sub-)Types (Abrupt Drift):} A small, distinct subset of the 41 CUAD clause types could be held out from early phases and introduced abruptly in a later phase.
        \end{enumerate}
\end{itemize}

\subsection{ARIA Agent Configuration}
\label{sec:aria_config}
The ARIA agent will be configured as follows:
\begin{itemize}[leftmargin=*,itemsep=0pt,parsep=0.2em,topsep=0.3em,partopsep=0.3em]
    \item \textbf{Internal LLM ($\ariallm$):} An LLM will be used for ARIA's internal reasoning, including initial prediction generation and self-assessment dialogues. The choice will be based on a balance of capability and inference cost/speed.
    \item \textbf{Initial Knowledge Repository ($\knowledgerepo_0$):} $\knowledgerepo_0$ will be initialized as empty or with a very small set of generic, high-level rules about contract language if found beneficial. The agent's parameters are $\params_i \approx \knowledgerepo_i$.
    \item \textbf{Decision Policy ($\pi(x_i; \knowledgerepo_i, \ariallm)$):}
        \begin{itemize}[leftmargin=*,itemsep=0pt,parsep=0.2em,topsep=0.3em,partopsep=0.3em]
            \item \textit{Clause Type Prediction:} $\ariallm$ is prompted with $x_i$ and a textual representation of the top-$k$ most relevant knowledge items retrieved from $\knowledgerepo_i$ (based on Temporally-Informed Knowledge Retrieval). The prompt will ask for the most likely clause type from the 41 CUAD categories and a confidence score.
        \end{itemize}
    \item \textbf{Intelligent Guidance Solicitation (IGS):}
        \begin{itemize}[leftmargin=*,itemsep=0pt,parsep=0.2em,topsep=0.3em,partopsep=0.3em]
            \item \textit{Self-Dialogue Reflective Questions ($RQ$):} Examples include:
            \begin{itemize}[leftmargin=*,itemsep=0pt,parsep=0.2em,topsep=0.3em,partopsep=0.3em]
                \item "Based on $x_i = \text{'...'}$, my predicted clause type is $\hat{y}_{\text{type}}$. What is my confidence (High/Moderate/Low) and why?"
                \item "Which specific phrases or keywords in $x_i$ support this classification? Are there any conflicting indicators?"
                \item "Retrieve relevant items from $\knowledgerepo_i$. How consistent is $\hat{y}_{\text{type}}$ with these items (e.g., definitions, exemplars, rules)?"
                \item "What are the key obligations and permissions implied by $x_i$ if it is indeed a $\hat{y}_{\text{type}}$?"
                \item "Is my knowledge regarding $\hat{y}_{\text{type}}$ (definitions, rules) in $\knowledgerepo_i$ marked as recently validated or potentially outdated?"
            \end{itemize}
            \item \textit{Confidence Self-Assessment ($conf_i \in \mathcal{C}$):} Based on the internal dialogue, $\ariallm$ will output a confidence level (e.g., High, Moderate, Low) for its clause type prediction.
            \item \textit{Query Decision ($d_i$):} If $conf_i \in \{\text{Moderate, Low}\}$ and the budget $B$ is not exhausted, ARIA sets $d_i = \text{query\_expert}$. The specific query $q_i \in \querytypes$ is chosen by IGS\_FormulateQuery based on the nature of the uncertainty identified in the self-dialogue (e.g., low confidence in type $\rightarrow q_{\text{type\_label}}$; uncertainty about implications $\rightarrow q_{\text{summarize\_rules}}$).
        \end{itemize}
    \item \textbf{Human-Guided Knowledge Adaptation (HGKA):}
        \begin{itemize}[leftmargin=*,itemsep=0pt,parsep=0.2em,topsep=0.3em,partopsep=0.3em]
            \item Oracle feedback $h_i$ (explanations, rules) will be parsed by $\ariallm$.
            \item \textit{Explanations:} Key phrases and concepts identified by the oracle will be stored as evidence linked to the clause $x_i$ (as an exemplar) and its true type $y_{i, \text{type}}^*$.
            \item \textit{Rule Summaries:} Oracle-provided rules will be canonicalized (e.g., into IF-THEN structures or semantic triples like `(ClauseType, has\_obligation, Action)`) and stored as new, validated knowledge items in $\knowledgerepo_i$. Each rule will have $kid, K, ts_{added}, ts_{validated}, S=\text{Valid}, M_{meta}$ (source=Oracle).
            \item \textit{Conflict Resolution:} If oracle feedback contradicts existing $KR$ items, the HGKA module will update status $S$ (e.g., to `PotentiallyOutdated` or `Superseded`) and timestamps, potentially triggering $q_{\text{clarify\_conflict}}$.
        \end{itemize}
\end{itemize}

\subsection{LLM-Simulated Human Expert Oracle ($\oraclellm$) Implementation}
\label{sec:llm_oracle}
The human expert oracle $\oracle$ will be simulated using a state-of-the-art LLM, denoted $\oraclellm$ (e.g., GPT-4 Turbo, Claude 3 Opus, Gemini 1.5 Pro). This $\oraclellm$ will be distinct from, and potentially more powerful or specifically prompted than, ARIA's internal $\ariallm$.

\begin{itemize}[leftmargin=*,itemsep=0pt,parsep=0.2em,topsep=0.3em,partopsep=0.3em]
    \item \textbf{General Persona Prompting:}
    Before processing specific queries, $\oraclellm$ will receive a system prompt like:
    \begin{quote}
    "You are an expert senior legal counsel specializing in commercial contract law. Your task is to provide precise, accurate, and actionable advice regarding contract clauses. When explaining clause types, clearly cite specific phrases from the provided clause text. When summarizing rules, define obligations and permissions for relevant parties. Adhere to current (simulated Phase [Phase Number]) legal best practices and interpretations."
    \end{quote}
    \item \textbf{Query-Specific Prompting for $\oraclellm$ to generate $h_i$:}
    \begin{description}
        \item[$q_{\text{type\_label}}(x_i)$:] The system will directly provide the ground truth $y_{i, \text{type}}^*$ from CUAD as $h_i$. The $\oraclellm$ is not used for this basic label query to ensure ground truth accuracy.
        \item[$q_{\text{explain\_type}}(x_i, y_{i, \text{type}}^*)$:]
            Prompt: "The following legal clause is classified as a '[value of $y_{i, \text{type}}^*$]'. Clause text: '[text of $x_i$]'. Please provide a concise explanation (2-3 sentences) for why this classification is correct, highlighting key phrases or legal concepts within the clause text that justify this type."
        \item[$q_{\text{summarize\_rules}}(x_i, y_{i, \text{type}}^*)$:]
            Prompt: "Consider the following legal clause, which is a '[value of $y_{i, \text{type}}^*$]': '[text of $x_i$]'. Summarize the key obligations, permissions, and significant provisions for the involved parties (use generic 'Party A' and 'Party B' if not specified) as a list of 2-4 short, structured rules. Example rule format: 'Party A MUST notify Party B within X days of event Y.'"
        \item[$q_{\text{clarify\_conflict}}(k_{\text{old}}, k_{\text{new}}, x_i)$:]
            Prompt: "Regarding the clause '[text of $x_i$]', my existing knowledge states: '[textual representation of $k_{\text{old}}$]'. However, new information suggests: '[textual representation of $k_{\text{new}}$]'. These appear to conflict. Please provide a clarification: Is one more accurate or relevant here? Should the old knowledge be updated or discarded? Explain your reasoning."
    \end{description}
    \item \textbf{Simulating Evolving Expertise (for Drift):} For different experimental phases (Section~\ref{sec:instantiation}), the system prompt for $\oraclellm$ or specific prompts for $q_{\text{summarize\_rules}}$ can be augmented with phase-specific instructions or references to (simulated) new legal precedents or policy changes. This makes the oracle's guidance itself dynamic.
\end{itemize}

\subsection{Baseline Models for Comparison}
\label{ssec:appendix_baselines}
ARIA's performance will be compared against several baselines, each configured as described below:
\begin{enumerate}[leftmargin=*,itemsep=0pt,parsep=0.2em,topsep=0.3em,partopsep=0.3em]
    \item \textbf{Static Base LLM (Zero-Shot/Few-Shot):} ARIA's internal $\ariallm$ is used to predict clause types for each instance $x_i$ based on a fixed prompt. This prompt may include a few representative examples of clauses and their types (few-shot) or no examples (zero-shot). No test-time learning, HITL interaction, or specialized $\knowledgerepo$ is used.
    \item \textbf{Static Fine-Tuned Model:} A smaller, efficient language model (e.g., a BERT-variant or a distilled LLM) is fine-tuned on a fixed initial portion of the CUAD stream (e.g., the first 10\% of instances, along with their true clause type labels). After fine-tuning, this model is applied without any further updates or adaptation to the rest of the stream.
\end{enumerate}

\end{document}